\documentclass{article}
\PassOptionsToPackage{numbers, compress}{natbib}

\usepackage[preprint]{style/neurips_2020}

\usepackage[utf8]{inputenc} 

\usepackage[T1]{fontenc}    
\usepackage{hyperref}       
\usepackage{url}            
\usepackage{booktabs}       
\usepackage{amsfonts}       
\usepackage{nicefrac}       
\usepackage{microtype}      

\usepackage{color}
\usepackage{amsmath}
\usepackage{amsthm}
\usepackage{bbm}
\usepackage{tikz}
\usepackage{mathdots}
\usepackage{yhmath}
\usepackage{cancel}
\usepackage{siunitx}
\usepackage{array}
\usepackage{multirow}
\usepackage{amssymb}
\usepackage{tabularx}
\usetikzlibrary{fadings}
\usetikzlibrary{patterns}
\usetikzlibrary{shadows.blur}

\usepackage{latexsym}
\usepackage{xspace}
\usepackage{listings}
\usepackage{enumitem}
\usepackage{balance}
\usepackage{algorithm}
\usepackage{mathtools}
\usepackage[position=bottom]{subfig}
\usepackage{float}
\usepackage{algpseudocode}

\newtheorem{definition}{Definition}
\newtheorem{proposition}{Proposition}

\newcommand*{\eat}[1]{}

\newcommand{\cfr}{counterfactually fair\xspace}

\newcommand{\igf}{in-group fairness\xspace}

\newcommand{\ratio}{IGF-Ratio\xspace}

\newcommand*{\rKL}{rKL\xspace}

\newcommand{\CM}{\textit{COMPAS}\xspace}
\newcommand{\MP}{\textit{MEPS}\xspace}

\newcommand{\MV}{\textit{moving company}\xspace}
\newcommand{\MVP}{\textit{moving company+BM}\xspace}
\newcommand{\MVR}{\textit{moving company+R}\xspace}

\newcommand{\WM}{White males\xspace}
\newcommand{\BM}{Black males\xspace}
\newcommand{\WF}{White females\xspace}
\newcommand{\BF}{Black females\xspace}

\def\btau{\boldsymbol{\tau}}
\def\bsigma{\boldsymbol{\sigma}}

\newcommand{\eg}{e.g.,\xspace}

\title{Causal intersectionality for fair ranking}

 \author{%
   Ke Yang \\
   New York University \\
   New York, NY \\
   \texttt{ky630@nyu.edu } \\
   \And
   Joshua R.~Loftus \\
   New York University \\
   New York, NY 10012 \\
   \texttt{loftus@nyu.edu} \\
  \And
   Julia Stoyanovich \\
   New York University \\
   New York, NY \\
  \texttt{stoyanovich@nyu.edu} \\
}

\begin{document}
\maketitle

\begin{abstract}
In this paper we propose a causal modeling approach to intersectional fairness, and a flexible, task-specific method for computing intersectionally fair rankings. Rankings are used in many contexts, ranging from Web search results to college admissions, but causal inference for fair rankings has received limited attention. Additionally, the growing literature on causal fairness has directed little attention to intersectionality. By bringing these issues together in a formal causal framework we make the application of intersectionality in fair machine learning explicit, connected to important real world effects and domain knowledge, and transparent about technical limitations. We experimentally evaluate our approach on real and synthetic datasets, exploring its behaviour under different structural assumptions.


\end{abstract}
\section{Introduction}
\label{sec:intro}

The machine learning community recognizes several important normative dimensions of information technology including privacy, transparency, and fairness. In this paper we focus on fairness --- a  broad and inherently interdisciplinary topic of which the social and philosophical foundations are still unresolved~\cite{DBLP:journals/cacm/ChouldechovaR20}.
To connect to these foundations, we take an approach based on \emph{causal modeling}. We assume that a suitable causal generative model is available and specifies relationships between variables including the \emph{sensitive attributes}, which define individual traits or social group memberships relevant for fairness. The model tells us how the world works, and we define fairness based on the model itself. In addition to being philosophically well-motivated, and to explicitly surfacing normative assumptions, the connection to causality gives us access to a growing literature on causal methods in general and causal fairness in particular.   

Research on fair machine learning has mainly focused on classification and prediction tasks, while we focus on ranking.
We consider two types of ranking tasks: score-based and learning to rank (LTR). In score-based ranking, a given set of candidates is sorted on the score attribute (which may itself be computed on the fly) and returned in sorted order.  In LTR, supervised learning is used to predict the ranking of unseen items.  In both cases, we typically return the highest scoring $k$ items, the top-$k$.  Set selection is a special case  of ranking that ignores the relative order among the top-$k$.

Further, previous research mostly considered a single sensitive attribute, while we use multiple sensitive attributes for the fairness component.  As noted by~\cite{crenshaw1989demarginalizing}, it is possible to give the appearance of being fair with respect to each sensitive attribute such as race and gender separately, while being unfair with respect to \emph{intersectional} subgroups. For example, if fairness is taken to mean proportional representation among the top-$k$, it is possible to achieve proportionality for each gender subgroup (e.g., men and women) and for each racial subgroup (e.g., Black and White), while still having inadequate representation for a subgroup defined by the intersection of both attributes (e.g., Black women). The literature on intersectionality includes numerous case studies, and theoretical and empirical work showing that people adversely impacted by more than one form of structural oppression face additional challenges in ways that are more than additive.

\begin{figure*}[t!]
	\centering
	\subfloat[original ranking]
	{\includegraphics[width=0.5\linewidth]{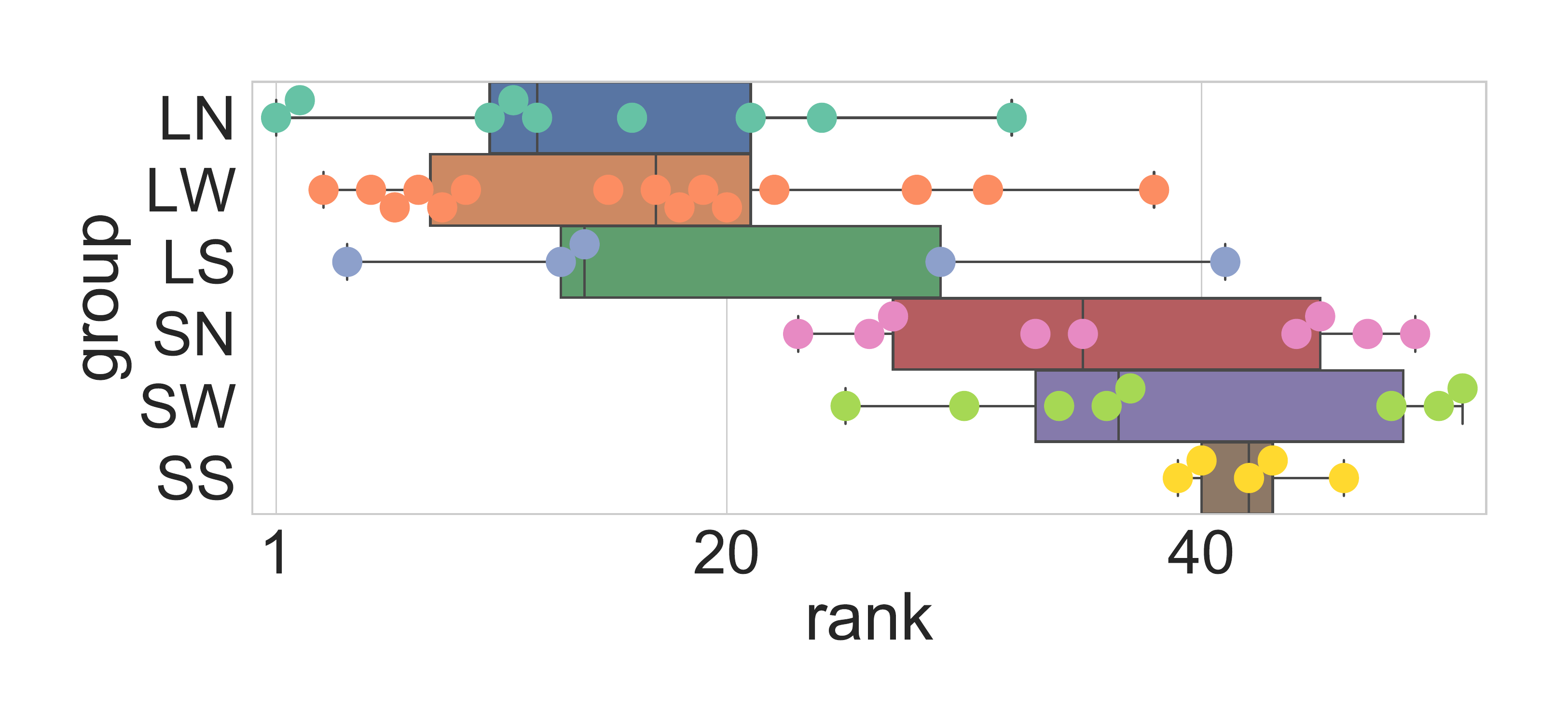}
	\label{fig:CS_rank_Y}
	}
	\subfloat[\cfr]{
	\includegraphics[width=0.5\linewidth]{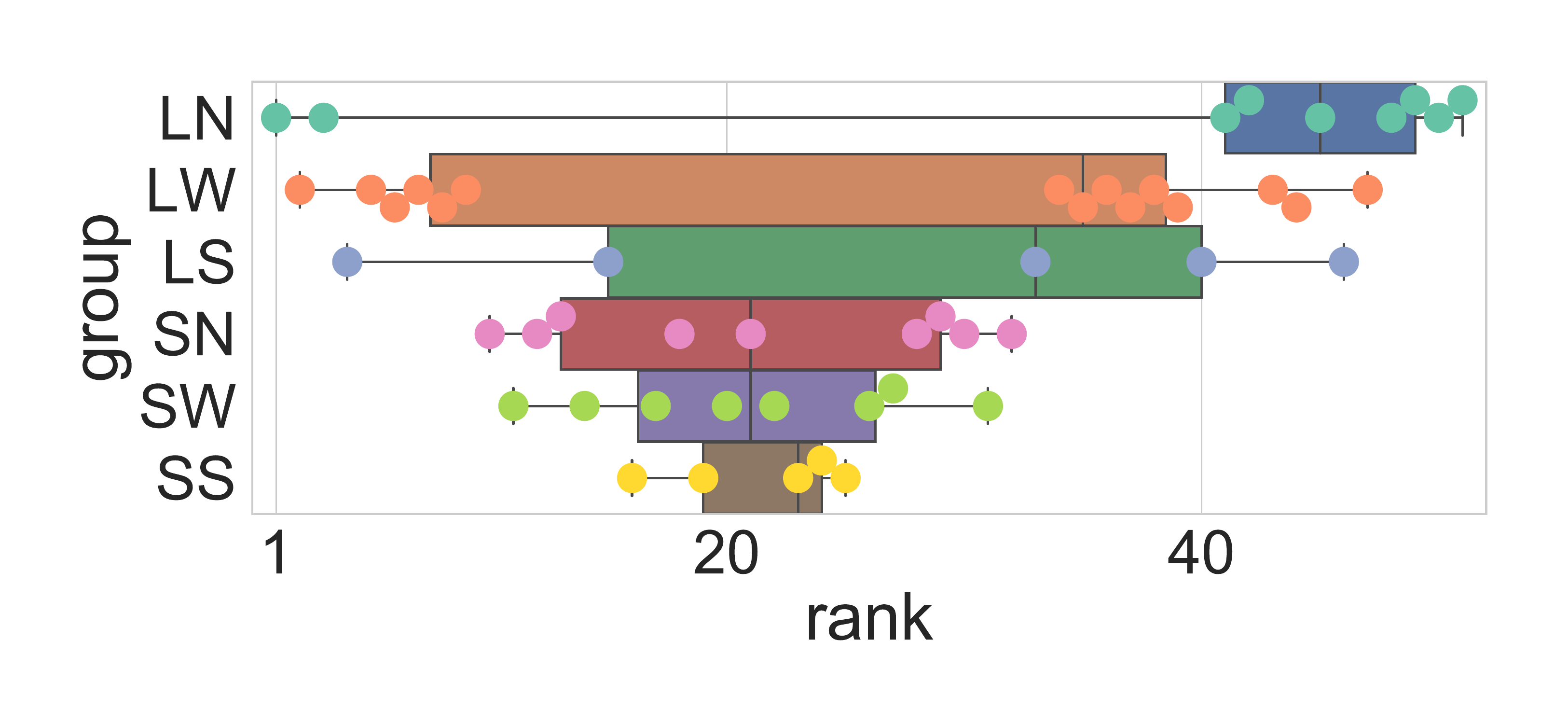}
	\label{fig:CS_rank_Y_count}
	}
	\caption{CSRanking by weighted publication count, showing positions of intersectional groups by department size, large (L) and small (S), and location, North East (N),  West (W), South East (S).}
	\label{fig:cs_rank}
\end{figure*}

With a given ranking task, set of sensitive attributes, and causal model, we propose ranking on counterfactual scores as a method to achieve intersectional fairness. From the causal model we compute model-based counterfactuals to answer the motivating question, ``What would this person's data look like if they had (or had not) been a Black woman (for example)?'' By ranking on counterfactual scores, \emph{we are treating every individual in the sample as though they had belonged to one specific intersectional subgroup}.
While intersectional concerns are usually raised when data is about people, they also apply for other types of entities.  Figure~\ref{fig:cs_rank} gives a preview of our method on the CSRankings dataset~\cite{csrankings} that ranks 51 CS departments in the US by a weighted publication count score (lower ranks are better).  Departments are of two sizes, large (L, with more than 30 faculty members) and small (S), and are located in three geographic areas.  The original ranking in Figure~\ref{fig:CS_rank_Y} prioritizes large departments, particularly in the North East and in the West.  The ranking in Figure~\ref{fig:CS_rank_Y_count} was derived using our method with size and location as sensitive attributes; it includes small departments at the top-20 and is more geographically balanced.

We define intersectional fairness for ranking in a similar manner to previous causal definitions of fairness for classification or prediction tasks~\cite{chiappa2019path, kilbertus2017avoiding, DBLP:conf/nips/KusnerLRS17, nabi2018fair, zhang2018fairness}. The idea is to model the causal effects of sensitive attributes on other variables and make algorithms fairer by removing these effects. Although the technical aspects of this approach are our main focus, we discuss some non-technical interpretation and limitation issues in the Broader Impact section. In the next section we introduce notation and describe the particular causal modeling approach we take, using directed acyclic graphs and structural equations, but we also note that our higher level ideas can be applied with other approaches to causal modeling. We experimentally demonstrate the effectiveness of our method on real and synthetic datasets in Section~\ref{sec:exp}.  We summarize related work in Section~\ref{sec:related} and conclude in Section~\ref{sec:conc}.  Our conclusions are supplemented by a discussion of the broader impacts of our work in Section~\ref{sec:bi}.  Our code is publicly available at \url{https://github.com/DataResponsibly/CIFRank}.
\section{Causal intersectionality}
\label{sec:fair}

In this section we describe the problem setting, our proposed definition of intersectional fairness within causal models, and an approach to computing rankings satisfying the fairness criterion.

\subsection{Model and problem setting}
\label{sec:fair_model}

\begin{figure*}[t!]
	\centering
	\subfloat[$\mathcal{M}_1$]{
		\includegraphics[width=0.24\linewidth]{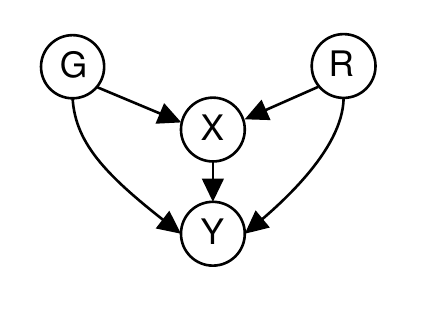}\label{fig:cm_covariate}
	}
	\subfloat[$\mathcal{M}_2$]{
		\includegraphics[width=0.24\linewidth]{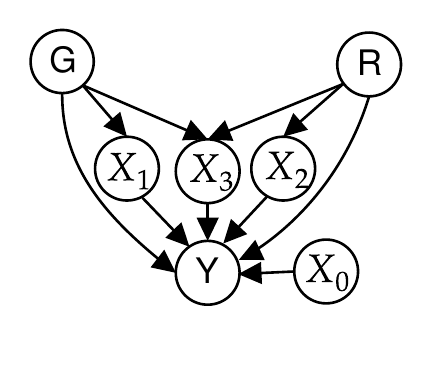}\label{fig:cm_three_m}
	}	
	\subfloat[$\mathcal{M}_3$]{\includegraphics[width=0.24\linewidth]{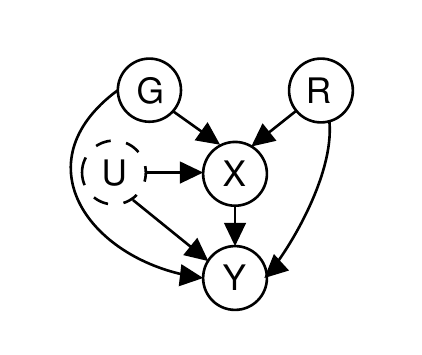}
		\label{fig:cm_covariate_unobserve}
	}
	\subfloat[$\mathcal{M}_4$]{
		\includegraphics[width=0.24\linewidth]{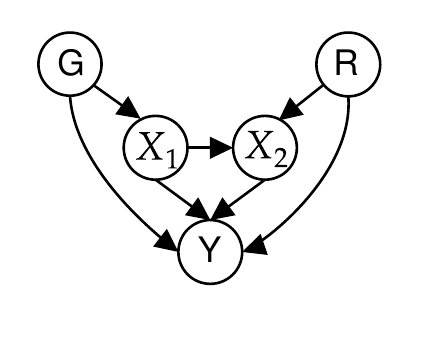}\label{fig:cm_two_m}
	}\hfill
	\caption{Causal models that include sensitive attributes $G$ (gender), $R$ (race), utility score $Y$, other covariates $\mathbf X$, and latent (unobserved) variable $U$.}
	\label{fig:cm}
\end{figure*}

\paragraph{Causal model}  As an input, our method requires a structural causal model (SCM), which we define briefly here and refer to~\cite{judea2000causality, loftus2018causal, robins1986new, spirtes2000causation} for more detail. An SCM consists of a directed acyclic graph (DAG), $\mathcal{G} = (\mathbf{V}, \mathbf{E})$, where the vertex set $\mathbf{V}$ represents variables, which may be observed or latent, and the edge set $\mathbf{E}$ indicates causal relationships from source vertices to target vertices. Several example DAGs are shown in Figure~\ref{fig:cm}, where vertices with dashed circles indicate latent variables.

For $V_j \in \mathbf{V}$ let $\text{pa}_j = \text{pa}(V_j) \subseteq \mathbf{V}$ be the ``parent'' set of all vertices with a directed edge into $V_j$. If $\text{pa}_j$ is empty we say $V_j$ is exogenous, and otherwise we assume that there is a function $f_j(\text{pa}_j)$ that approximates the expectation or some other link function, such as the log-odds, of $V_j$. Depending on background knowledge or the level of assumptions we are willing to hazard, we assume the functions $f_j$ are either known or can be suitably estimated from the data. We also assume a set of sensitive attributes $\mathbf A \subseteq \mathbf V$, chosen \emph{a priori}, for which existing legal, ethical, or social norms suggest that the ranking algorithm should be fair.

\paragraph{Problem setting}  In most of our examples we consider two sensitive attributes, which we denote $G$ and $R$ motivated by the example~\cite{crenshaw1989demarginalizing} of gender and race. We let $Y$ denote an outcome variable that is used as a utility score in our ranking task, and $\mathbf X$ be \emph{a priori} non-sensitive predictor variables. In examples with pathways from sensitive attributes to $Y$ passing through $\mathbf X$ we call the affected variables in $\mathbf X$ mediators. Finally, $U$ may denote an unobserved confounder. In some settings a mediator may be considered \emph{a priori} to be a legitimate basis for decisions even if it results in disparities. This is what~\cite{DBLP:journals/corr/abs-1807-08362} calls the infra-marginality principle, \cite{chiappa2019path, DBLP:conf/nips/KusnerLRS17, nabi2018fair} refer to as path-specific effects and \cite{zhang2018fairness} as indirect effects, and~\cite{kilbertus2017avoiding} calls such mediators \textbf{resolving variables}. We adopt the latter terminology and will show examples of different cases later. In fact, our method allows mediators to be resolving for one sensitive attribute and not the other, reflecting  nuances that may be necessary in intersectional problems. 

For simplicity of presentation we treat sensitive attributes as binary indicators of a particular privileged status rather than a more fine grained coding of identity, but this is not a necessary limitation of the method. Our experiments in Section~\ref{sec:exp} use model $\mathcal{M}_1$ in Figure~\ref{fig:cm_covariate}, but richer datasets and complex scenarios as in Figure~\ref{fig:cm_three_m} fit into our framework.
\emph{Sequential ignorability}~\cite{imai2010identification, pearl2001direct, robins2003semantics, vanderweele2015explanation} is a standard assumption for model identifiability that can be violated by unobserved confounding between a mediator and outcome as in $\mathcal{M}_3$, or observed confounding where one mediator is a cause of another as in $\mathcal{M}_4$ in the Figure. We include these as indications of qualitative limitations of this framework.

\subsection{Counterfactual intersectional fairness}

\paragraph{Intersectionality}  It is common in predictive modeling to assume a function class that is linear or additive in the inputs, that is, for a given non-sensitive variable $V_j$ 

\begin{equation*}
    f_j(\text{pa}_j) = \sum_{V_l \in \text{pa}_j} f_{j,l}(V_l).
\end{equation*}

Such simple models may be less likely to overfit and are more interpretable. However, to model the intersectional effect of multiple sensitive attributes \emph{we must avoid this assumption}. Instead, we generally assume that $f_j$ contains non-additive interactions between sensitive attributes. With rich enough data such non-linear $f_j$ could be modeled flexibly, but to keep some simplicity in our examples we will consider functions with linear main effects and second order interactions. That is, if $\text{pa}_j$ includes sensitive attributes $A_1, A_2, \ldots$, and non-sensitive attributes $V_1, V_2, \ldots$, we assume

\begin{equation}
    f_j(\text{pa}_j) = \sum_l \beta_{j,l} V_l + \sum_{l} \eta_{j,l} A_l + \sum_{r} \sum_{l \neq r} \eta_{j,r,l} A_r A_l.
    \label{eq:scm}
\end{equation}

Since our motivating examples do not involve numerical sensitive attributes we do not pursue higher order interactions, but that can be done in the current framework. The coefficients (or weights) $\eta_{j,l}$ model the (marginal) causal effect on $V_j$ of structural oppression on the basis of sensitive attribute $A_l$, while $\eta_{j,r,l}$ model the non-additive combination of adversity due to the multiple intersecting forms of structural oppression related to both $A_r$ and $A_l$.

Our experiments use simpler examples with one mediator and binary sensitive attributes so the results are easier to interpret and compare to non-causal notions of fairness in ranking. Non-binary cases and sophisticated models like Figure~\ref{fig:cm_three_m} with combinations of resolving and non-resolving mediators would be more difficult to compare to other approaches, but we believe this reflects that real-world intersectionality can pose hard problems that our framework is capable of analyzing. And while identifiability and estimation are simplified in binary examples, the growing literature on causal mediation discussed in Section~\ref{sec:related} can be used on harder problems.

\paragraph{Counterfactuals} Causal models allow us to compute model-based counterfactuals, which we interpret informally as ``the value $Y$ would have taken if $G$ or $R$ had been different than their actual values.'' Letting $\mathbf A$ denote the vector of sensitive attributes and $\mathbf a$ a possible value for these, we compute the counterfactual $Y_{\mathbf A \gets \mathbf a'}$ by replacing the observed value of $\mathbf A$ with $\mathbf a'$ and then propagating this change through the DAG: any directed descendant $V_j$ of $\mathbf A$ has its value changed by computing $f_j(\text{pa}_j)$ with the new value of $\mathbf a'$, and this operation is iterated until it reaches all the terminal nodes that are descendants of any of the sensitive attributes $\mathbf A$. This process is well-defined for the graphs we consider because there are no directed cycles and the sensitive attributes are assumed to be exogenous. For graphs with resolving mediators we modify this process as follows. We provide more detail specifically for model $\mathcal{M}_1$ in Figure~\ref{fig:cm_covariate} with both the resolving and non-resolving cases. We focus on this model for clarity, but all that we say in the rest of this section requires only minor changes to hold for other models such as $\mathcal M_2$ without loss of generality, provided they satisfy sequential ignorability. Our implementation is similar to what ~\cite{DBLP:conf/nips/KusnerLRS17} refer to as "Level 3" assumptions, but we denote exogenous error terms as $\epsilon$ instead of $U$.

The model includes sensitive attributes $G$ and $R$, (continuous) mediator $X$, and outcome $Y$ satisfying

\begin{equation*}
    X = f_X(G, R) + \epsilon^X, \quad  Y = f_Y(X, G, R) + \epsilon^Y.
\end{equation*}

We mention in passing that the case where $Y$ is not continuous fits in the present framework with minor modifications, where we have instead a probability model with corresponding link function $g$ so that

\begin{equation*}
    \mathbb E[Y|X,G,R] = g^{-1}(f_Y(X, G, R))
\end{equation*}

Suppose the observed values for observation $i$ are $(y_i, x_i, g_i, r_i)$ with exogenous errors $\epsilon^X_i, \epsilon^Y_i$. Since we do not model any unobserved confounders in model $\mathcal M_1$, we suppress the notation for $U$ and denote counterfactual scores, for some $(g',r') \neq (g,r)$, as

\begin{equation*}
    Y_i' := (Y_i)_{\mathbf A \gets \mathbf a'} = (Y_i)_{(G, R) \gets (g', r')}.
\end{equation*}

If $X$ is \textbf{non-resolving} then we first compute counterfactual $X$ as $x_i' := f_X(g', r') + \epsilon^X_i$, substituting $(g', r')$ in place of the observed $(g_i, r_i)$. Then we do the same substitution while computing

\begin{equation*}
    Y_i' = f_Y(x_i', g', r') + \epsilon^Y_i = f_Y(f_X(g',r') + \epsilon^X_i, g', r') + \epsilon^Y_i.
\end{equation*}

If $X$ is \textbf{resolving} then we keep the observed $X$ and compute

\begin{equation*}
    Y_i' = f_Y(x_i, g', r') + \epsilon^Y_i.
\end{equation*}

If the functions $f_X, f_Y$ have been estimated from the data then we have observed residuals $r^X_i, r^Y_i$ instead of model errors in the above.

\subsection{Counterfactually Fair Ranking}

\paragraph{Ranking task}  We use an outcome or utility score $Y$ to rank a dataset $\mathbf{D}$ assumed to be generated by a model $\mathcal{M}$ from among the example SCMs in Figure~\ref{fig:cm}. If the data contains a mediating predictor variable $X$ then the task also requires specification of whether or not $X$ is a resolving mediator. Letting $n = |\mathbf{D}|$, a ranking is a permutation $\btau = \hat \btau(\mathbf{D})$ of the $n$ individuals or items, usually satisfying

\begin{equation}
    Y_{\btau(1)} \geq Y_{\btau(2)} \geq \cdots \geq Y_{\btau(n)}.
\end{equation}

To satisfy other objectives, like fairness, we generally output a ranking $\hat \btau$ that is not simply sorting on the observed values of $Y$. Specifically, we aim to compute rankings satisfying

\begin{definition}[Counterfactually fair ranking]
A ranking $\hat \btau$ is counterfactually fair if, for all possible $x$ and pairs of vectors of actual and counterfactual sensitive attributes $a \neq a'$, respectively, we have
\begin{equation}
\label{eq:cfranking}
    \mathbb P (\hat \btau(Y_{\mathbf A \gets \mathbf a}(U)) = k \mid \mathbf X = \mathbf x, \mathbf A = \mathbf a) = \mathbb P (\hat \btau(Y_{\mathbf A \gets \mathbf a'}(U)) = k \mid \mathbf X = \mathbf x, \mathbf A = \mathbf a)
\end{equation}
for any rank $k$, and with suitably randomized tie-breaking.
\end{definition}

We propose this definition as a natural adaptation of causal definitions in~\cite{chiappa2019path, kilbertus2017avoiding, DBLP:conf/nips/KusnerLRS17, nabi2018fair, zhang2018fairness} for fairness in classification and prediction tasks to the ranking setting. To satisfy Equation~\ref{eq:cfranking} we propose ranking using counterfactuals that treat all individuals or items in the dataset as though they have a certain intersecting set of privileges or disadvantages. That is, we fix a reference subgroup using one value of $a$ and transform the data so that each observation has this value of $a$, and the consequences thereof, as computed from the DAG.

\paragraph{Implementation} 
We use the following procedure to compute counterfactually fair rankings.

\begin{enumerate}
    \item For a (training) dataset $\mathbf{D}$ we estimate the parameters of the assumed causal model $\mathcal{M}$. A variety of frequentist or Bayesian approaches for estimation can be used. Our experiments
    use the R package \texttt{mediation}~\cite{tingley2014mediation} on model $\mathcal{M}_1$ in Figure~\ref{fig:cm_covariate}. 
    \item From the estimated causal model we compute counterfactual records on the (training) data, transforming each observation to one reference subgroup $\mathbf A \gets \mathbf a'$, we set $\mathbf a'$ to be the unprivileged intersectional group. This yields counterfactual training data $\mathbf{D}_{\mathbf A \gets \mathbf a'}$.
    \item For score based ranking we sort $Y_{\mathbf A \gets \mathbf a'}$ in descending order to produce the \cfr ranking $\hat \btau(Y_{\mathbf A \gets \mathbf a'})$. For learning to rank (LTR), we apply a learning algorithm on $\mathbf{D}_{\mathbf A \gets \mathbf a'}$ and consider two options depending on whether the problem structure allows use of the causal model at test time: if it does then we in-process the test data from the learned causal model before ranking counterfactual test scores, and if not we rank the unmodified test data. We refer to the first case as cf-LTR and emphasize that in the second case \emph{counterfactually fairness may not hold, or hold only approximately, on test data}.
\end{enumerate}

Next we show that this implementation, under common causal modeling assumptions, satisfies our fair ranking criteria.

\begin{proposition}[Implementing counterfactually fair ranking]
If the assumed causal model $\mathcal{M}$ is identifiable and correctly specified, implementations described above produce counterfactually fair rankings in the score based ranking and cf-LTR tasks. 
\end{proposition}

The proof is essentially by construction, but we provide more detail now for model $\mathcal{M}_1$. Fixing a baseline intersectional subgroup $(g_0, r_0)$, the counterfactual training data in our implementation will use

\begin{equation*}
    Y_{(G,R) \gets (g_0, r_0)},
\end{equation*}

either by ranking these for score based ranking or training a predictive model for LTR. We wish to show that

\begin{equation}
    \label{eq:cfr}
     \mathbb P (\hat \btau(Y_{(G,R) \gets (g,r)}) = k \mid X = x, (G,R) = (g,r)) 
\end{equation}

is unchanged under all counterfactual transformations $Y_{(G,R) \gets (g',r')}$ if the causal model has been correctly specified.

\begin{proof}[\textbf{Proof of Proposition 1}]
First, we consider the case where the functions $f_X, f_Y$ are known. If $X$ is resolving, then

\begin{equation*}
    (Y_i)_{(G,R) \gets (g_0, r_0)} = f_Y(x_i, g_0, r_0) + \epsilon^Y_i
\end{equation*}

for all $i$. In this case the conditional distribution of these scores~\eqref{eq:cfr} is invariant under counterfactual transformations $(g,r) \gets (g',r')$ because $x_i$ is held fixed, $(g', r')$ will be substituted with the fixed baseline values $(g_0, r_0)$, and the error term is exogenous and in particular its distribution does not change under transformations of $(g,r)$. If $X$ is not resolving then we use

\begin{equation*}
    (Y_i)_{(G,R) \gets (g_0, r_0)} = f_Y(f_X(g_0,r_0) + \epsilon^X_i, g_0, r_0) + \epsilon^Y_i
\end{equation*}

Under counterfactual transformations $(g,r) \gets (g',r')$ all of the inputs above stay fixed except for the error terms, and, as before, these errors do not depend on $(g,r)$ so the training data scores have the desired distributional invariance. 

For score based ranking $\hat \btau$ sorts the counterfactual scores $(Y_i)_{(G,R) \gets (g_0, r_0)}$. Since the distributions of these scores are unchanged under counterfactual transformations, the probability for any score to equal a given rank $k$ is also unchanged, hence $\hat \btau$ is a counterfactually fair ranking.

In cf-LTR, at test time the test data is first transformed to $\mathbf D^{\text{test}}_{(G,R) \gets (g_0, r_0)}$ before inputting to $\hat \btau$.  As before, the distribution of the predicted rank for observation $i$ under any counterfactual transformation ${(G,R) \gets (g', r')}$ is fixed to that of the distribution under ${(G,R) \gets (g_0, r_0)}$, which depends only on the exogenous errors.

Finally, we relax the assumption that the functions $f_X, f_Y$ are known. Since we have assumed the causal model is identifiable and correctly specified (in particular, it satisfies sequential ignorability in cases where the model has mediators), these functions can be estimated on the (training) data via any appropriate causal inference method. Hence, counterfactually fair ranking condition will hold approximately due to plug-in estimation error. 
\end{proof}

\section{Experimental Evaluation}
\label{sec:exp}

In this section we investigate the behavior of our framework under different structural assumptions of the underlying causal model on real and synthetic datasets.  We quantify performance with respect to several fairness and utility measures, for both score-based rankers and for learning to rank (LTR). 

\subsection{Datasets and Evaluation Measures}
\label{sec:exp:methods}

\paragraph{Datasets} We present experimental results on two real datasets, \CM~\cite{angwin2016machine} and \MP~\cite{cohen2009medical}, and on a synthetic benchmark that simulates hiring by a moving company, inspired by Datta et al.~\cite{DBLP:conf/sp/DattaSZ16}.  

\begin{itemize}
\item \CM contains arrest records with sensitive attributes gender and race.  We use a subset of this data that includes Black and White individuals of either gender with at least 1 prior arrest.  The resulting dataset has 4,162 records with about 25\% \WM, 59\% \BM, 6\% \WF, and 10\% \BF.  We fit  the causal model $\mathcal{M}_1$ in Figure~\ref{fig:cm_covariate} with gender $G$, race $R$, number of prior arrests $X$, and COMPAS decile score $Y$, with larger $Y$ predicting higher likelihood of recidivism.

\item \emph{Medical Expenditure Panel Survey (\MP)} is a comprehensive source of individual and household-level information regarding the amount of health expenditures by individuals from various demographic or socioeconomic groups.  We compute $Y$ as the sum of the number of  office-based, outpatient, home health, and ER visits, and the number of inpatient nights, as is done in the IBM AIF 360 toolkit~\cite{bellamy2018ai}.  The resulting dataset has 15,675 records with about 18\% \WM, 29\% \BM, 20\% \WF, and 33\% \BF. We fit a causal model $\mathcal{M}$, in which gender $G$, race $R$, and age $X$ causally affect the utilization score $Y$. This is similar to $\mathcal{M}_2$ in Figure 2b with $X_0 = X$ but no mediators. Age $X$ here is a moderator that we flexibly model using natural cubic splines \emph{within each intersectional group}. We use this dataset to illustrate the flexibility of our framework to include nonlinear interactions with continuous variables. Note that a higher value of $Y$ indicates higher  medical resource utilization.  We target the use case where ranking on $Y$ from higher to lower is used to address the triage problem, helping prioritize individuals for additional assistance.

\item \MV is a synthetic dataset drawn from the causal model $\mathcal{M}_1$ in Figure~\ref{fig:cm_covariate}, with edge weights: $w(G \rightarrow X)=1$, $w(R \rightarrow X) = 0$, $w(G \rightarrow Y) = 0.12$, $w(R \rightarrow Y) = 0.08$, and $w(X \rightarrow Y) = 0.8$. Job applicants are hired by the moving company based on their qualification score $Y$, computed from weight-lifting ability $X$, and affected by gender $G$ and race $R$, either directly or through $X$.  Specifically, weight-lifting ability $X$ is lower for female applicants than for male applicants; qualification score $Y$ is lower for female applicants and for Blacks.  Thus, the intersectional group \BF faces greater discrimination than either the Black or the female group.  In our experiments in this section, we assume that women and Blacks constitute a 37\% minority of the applicants, and that gender and race are assigned independently.  As a result, there are about 40\% \WM, 14\% \BF, and 23\% of both \BM and \WF in the input. 

\item \MVR is a synthetic dataset drawn from the causal model $\mathcal{M}_1$ in Figure~\ref{fig:cm_covariate}, with the same proportion of intersectional groups as in \MV, but with edge weight  $w(R \rightarrow X)=0.1$. Other edge weights in \MVR are the same as in \MV above. In other words, race impacts the outcome $Y$ both directly and indirectly through $X$ in \MVR, while in \MV we assumed a direct effect only.  
Black applicants have greater weight-lifting ability, perhaps due to more previous experience in similar roles, but we emphasize this is only a hypothetical scenario not meant to reflect any real world beliefs. The direct effect of race on $Y$ is negative due to discrimination but the mediated effect through $X$ is positive, so the total observed effect is smaller.  We use this dataset to illustrate the behavior of our framework under different assumptions about the impact of the sensitive attributes on the mediator and outcome, and show different behavior for the resolving and non-resolving cases. 

\item \MVP is a synthetic dataset drawn from the causal model $\mathcal{M}_1$ in Figure~\ref{fig:cm_covariate}, with the same edge weights as specified in \MV, but with a different proportion of intersectional groups: there are more \BM in \MVP compared to \MV. There are about 50\% \BM, 40\% \WM, 5\% \WF, and 5\% \BF. We use this dataset to illustrate the behavior of our framework for different  compositions of the dataset. 
\end{itemize}

\paragraph{Fairness measures} We quantify performance using two well-understood fairness measures that also have a natural interpretation for rankings: {\bf demographic parity at top-$k$} and {\bf equal opportunity at top-$k$}, for varying values of $k$. We also use two rank-aware fairness measures, {\bf normalized discounted KL-divergence (\rKL)} and {\bf in-group fairness ratio (\ratio)}.

\begin{itemize}

\item Demographic parity (DP) is achieved if the proportion of the individuals belonging to a particular group corresponds to their proportion in the input.  We will represent DP by showing selection rates for each intersectional group at the top-$k$, with a value of $1$ for all groups corresponding to perfect DP.  

\item Equal opportunity (EO) in binary classification is achieved when the likelihood of  receiving the positive prediction for items whose true label is positive does not depend on the values of their sensitive attributes~\cite{DBLP:conf/nips/HardtPNS16}.  To measure EO for LTR, we will take the set of items placed at the top-$k$ in the ground-truth ranking to correspond to the positive class \emph{for that value of $k$}.  We will then present sensitivity (true positives / true positives + false negatives) per intersectional group at the top-$k$.  If sensitivity is equal for all groups, then the method achieves EO.

\item Normalized discounted KL-divergence (\rKL) was proposed in~\cite{DBLP:conf/ssdbm/YangS17} and is based on the intuition that, because it is more beneficial for an item to be ranked higher, it is also more important to achieve fairness (as measured, for example, by demographic parity or by ratio difference) at higher ranks.  This rank-awareness is achieved by taking a set-wise measure and placing it within the nDCG framework, which provides a principled rank-based discounting mechanism.   \rKL is one of three rank-aware fairness measures proposed in~\cite{DBLP:conf/ssdbm/YangS17}; all of them are designed for binary protected group membership.  We choose \rKL here because it is less sensitive to an imbalance in proportions of group sizes in the input, and because it can be extended to handle non-binary protected group membership.  For a given ranking $\btau$ of lendth $N$, we compute:

\begin{equation*}
\rKL(\btau)= \sum_{i=10,20,...}^{N}{ \frac{D_{KL}(P_i||Q)}{log_{2}{i}}}
\end{equation*}

Here, $D_{KL}(P_i||Q)$ is the Kullback-Leibler divergence between two discrete probability distributions $P_i$ and $Q$, with $P_i(g)$ set to the proportion of group $g$ at the top-$i$ and $Q(g)$ set to the proportion of $g$ in the input.

\item In-group fairness ratio (\ratio) is the simpler of two in-group fairness measures proposed in~\cite{DBLP:conf/ijcai/YangGS19}.  It  capture an important intersectional concern that arises when an input ranking must be re-ordered (and thus suffer a utility loss) to satisfy some fairness or diversity constraint.  Specifically, \ratio compares the amount of re-ordering within each intersectional groups, and considers a ranking fair if the corresponding loss is balanced across groups.  

Recall that $\btau_{1\ldots k}$ denotes the \emph{set} of the top-$k$ items in a ranking $\btau$.  For a given intersectional group $g$ and position $k$, $\textit{IGF-Ratio}_k(\btau,g)$ is the ratio of lowest score of any item from $g$ in $\btau_{1\ldots k}$ and the highest score of an item from $g$ not in $\btau_{1\ldots k}$.  \ratio works for non-negative scores and ranges from $[0, 1]$, with higher values implying better \igf.  To make the scores generated by our framework non-negative, we increase the values of $Y$ by $|\text{min}(Y)|$. 
\end{itemize}

\paragraph{Utility measures} We use two utility measures, {\bf $Y$-utility loss at top-$k$}, applicable to both score-based and learned rankers, and {\bf average precision}, applicable to learned rankers. Both compare a "ground truth" ranking $\btau$ induced by the order of the observed scores $Y$ to a proposed ranking $\bsigma$ which attempts to be fair, for example $\bsigma$ may be based on quota system taking the top ranked observations in each subgroup or could be a counterfactually fair $\hat \btau$ computed using our method.

\begin{itemize}

\item $Y$-utility loss at top-$k$ is defined as $L_k(\bsigma) = 1 - \sum_{i=1}^{k} Y_{\bsigma(i)} / \sum_{i=1}^{k} Y_{\btau(i)}$, where $Y_{\bsigma(i)}$ is the observed score of the item that appears at position $i$ in $\bsigma$ and $Y_{\btau(i)}$ is the observed score at rank $i$ in the original order $\btau$.  $L_k$ ranges between 0 (best) and 1 (worst).  

\item Average precision (AP) quantifies, in a rank-compounded manner, how many of the items that should be returned among the top-$k$ are indeed returned.   Let us denote by $\btau_{1\ldots k}$ the \emph{set} of the top-$k$ items in $\btau$.  We define precision at top-$k$ as $P_k = |\btau_{1\ldots k} \cap \bsigma_{1\ldots k}| / k$, where $\btau$ is the ground truth ranking and $\bsigma$ is the predicted ranking.  Then, $AP_k(\bsigma) = \sum_{i=1}^k P_i \times \mathbbm{1}{[\bsigma(i) \in \btau_{1\ldots k}]} / k$, where $\mathbbm{1}$ is an indicator function that returns 1 if the condition is met and 0 otherwise.  $AP_k$ ranges between 0 (worst) and 1 (best).
\end{itemize}

\begin{figure*}[t!]
	\centering
	\subfloat[original ranking]{\includegraphics[width=0.33\linewidth]{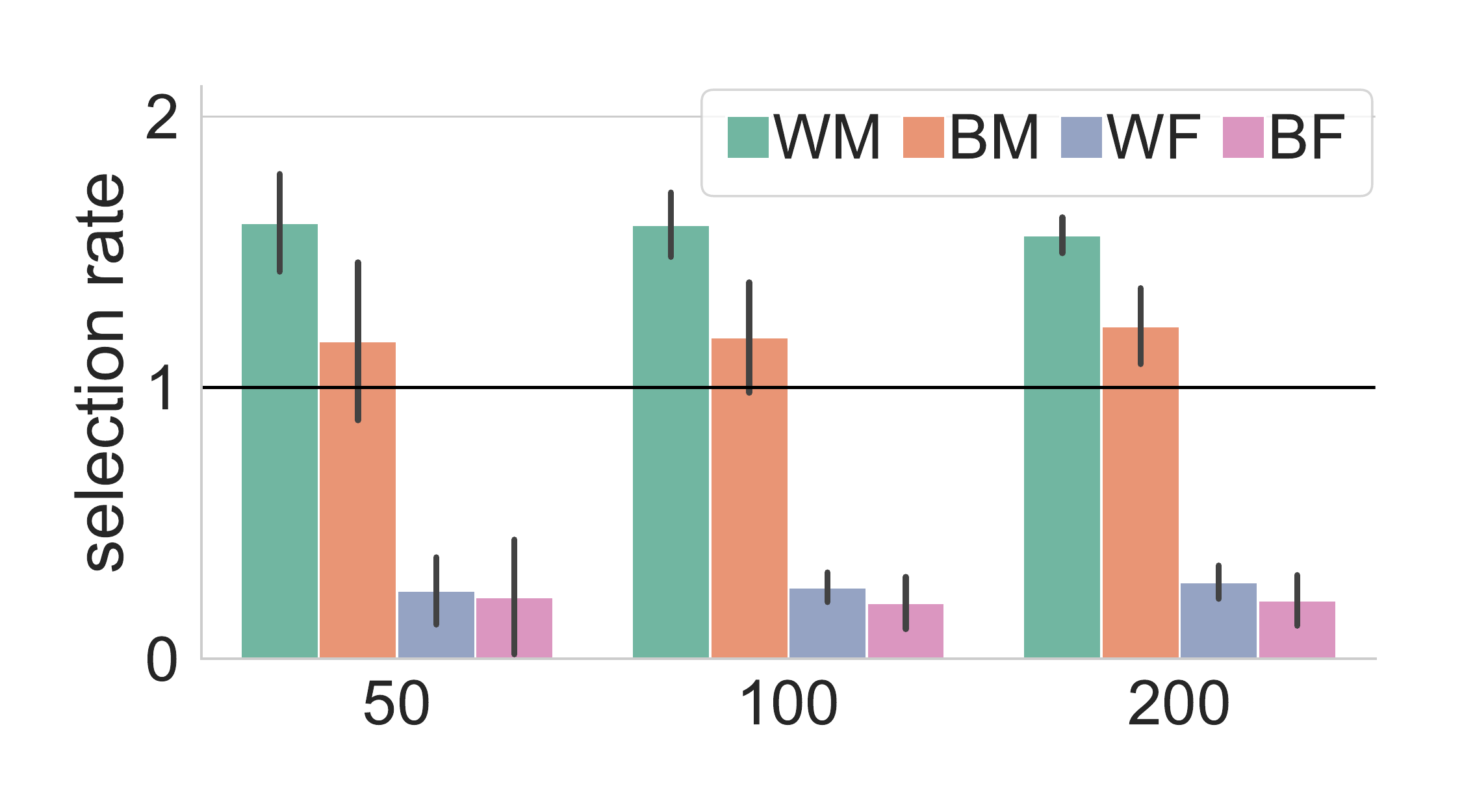}
	\label{fig:mv_dp_Y}
	}
	\subfloat[\cfr]{
	\includegraphics[width=0.33\linewidth]{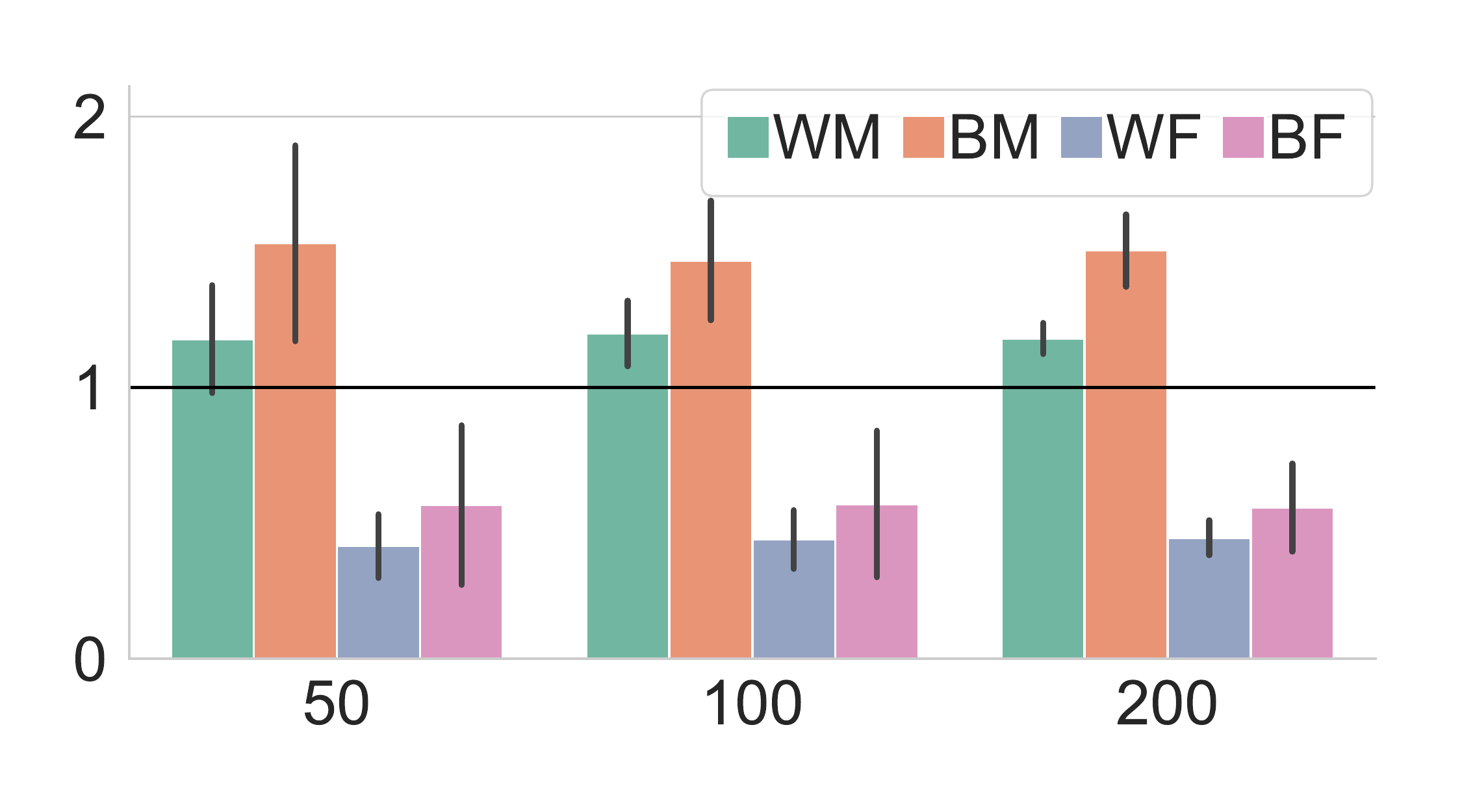}\label{fig:mv_dp_Y_count_resolve}
	}
	\subfloat[quotas on $R$]{
	\includegraphics[width=0.33\linewidth]{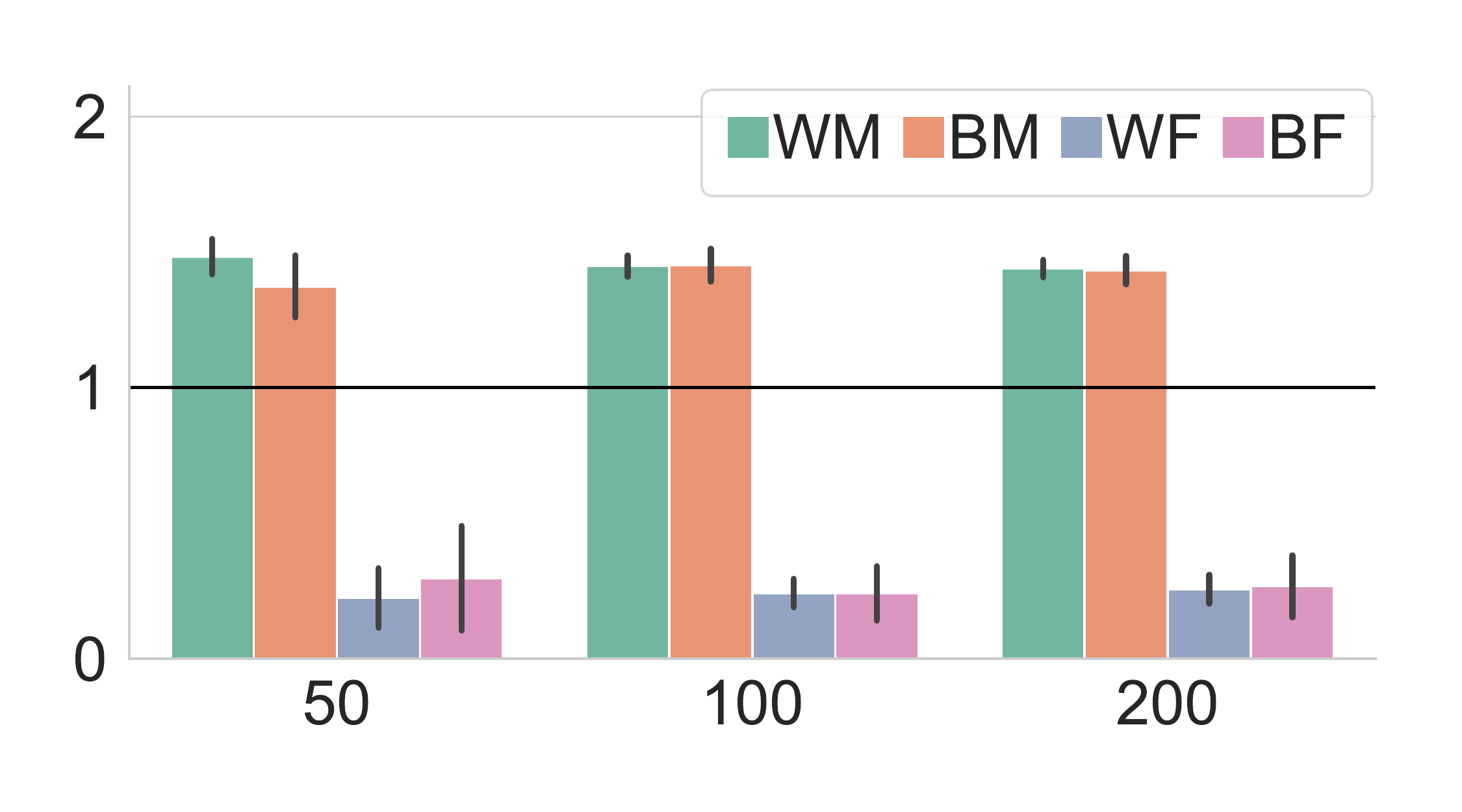}\label{fig:mv_dp_Y_quotas_R}
	}
	\caption{Demographic parity on the \MV dataset, generated using causal model $\mathcal{M}_1$ in Figure~\ref{fig:cm_covariate} with gender $G$, race $R$, resolving mediator weight-lifting ability $X$, and qualification $Y$.}
	\label{fig:mv_dp}
\end{figure*}
\begin{figure*}[h!]
	\centering
	\subfloat[original ranking]{\includegraphics[width=0.33\linewidth]{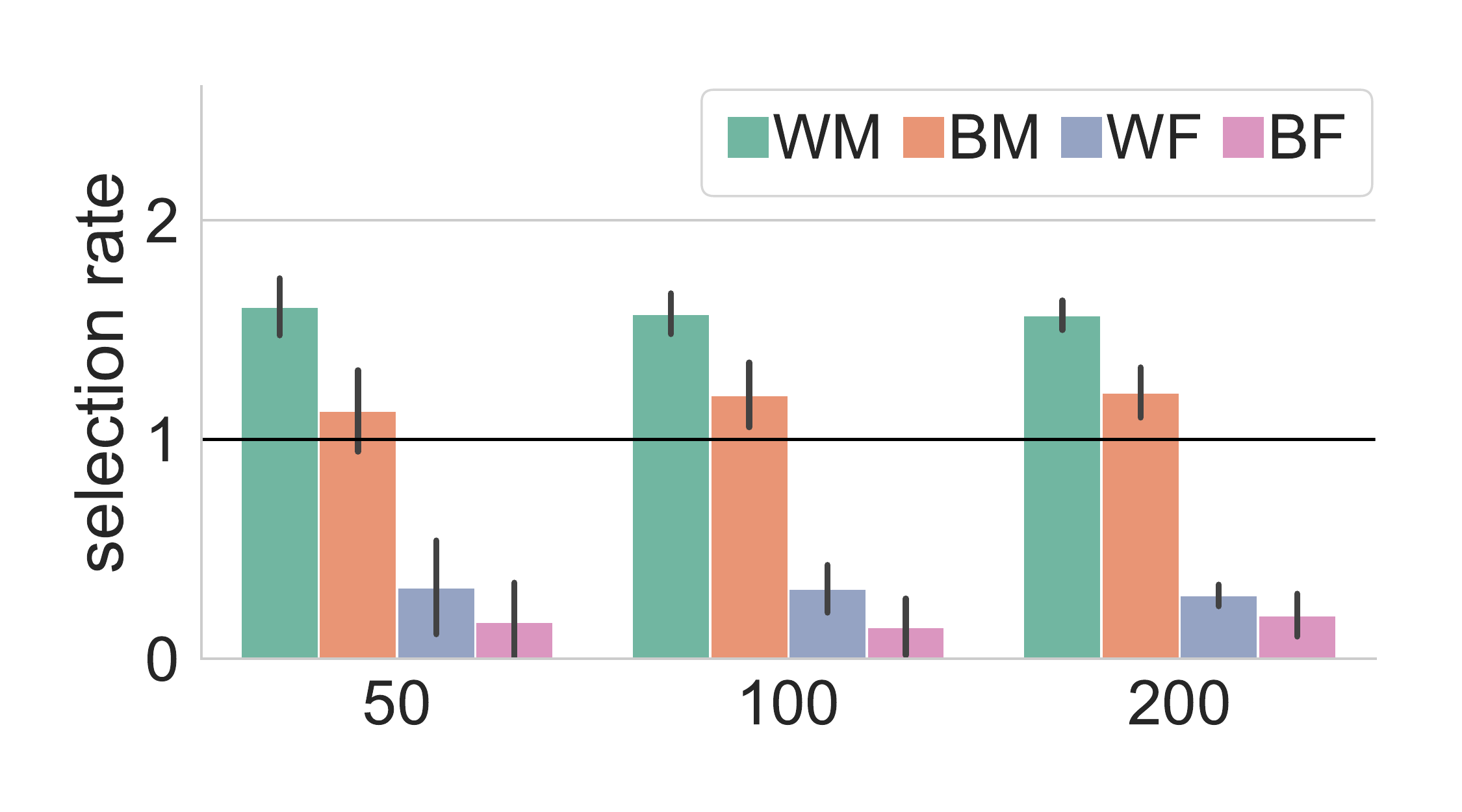}
	\label{fig:mvr_dp_Y}
	}
	\subfloat[\cfr]{
	\includegraphics[width=0.33\linewidth]{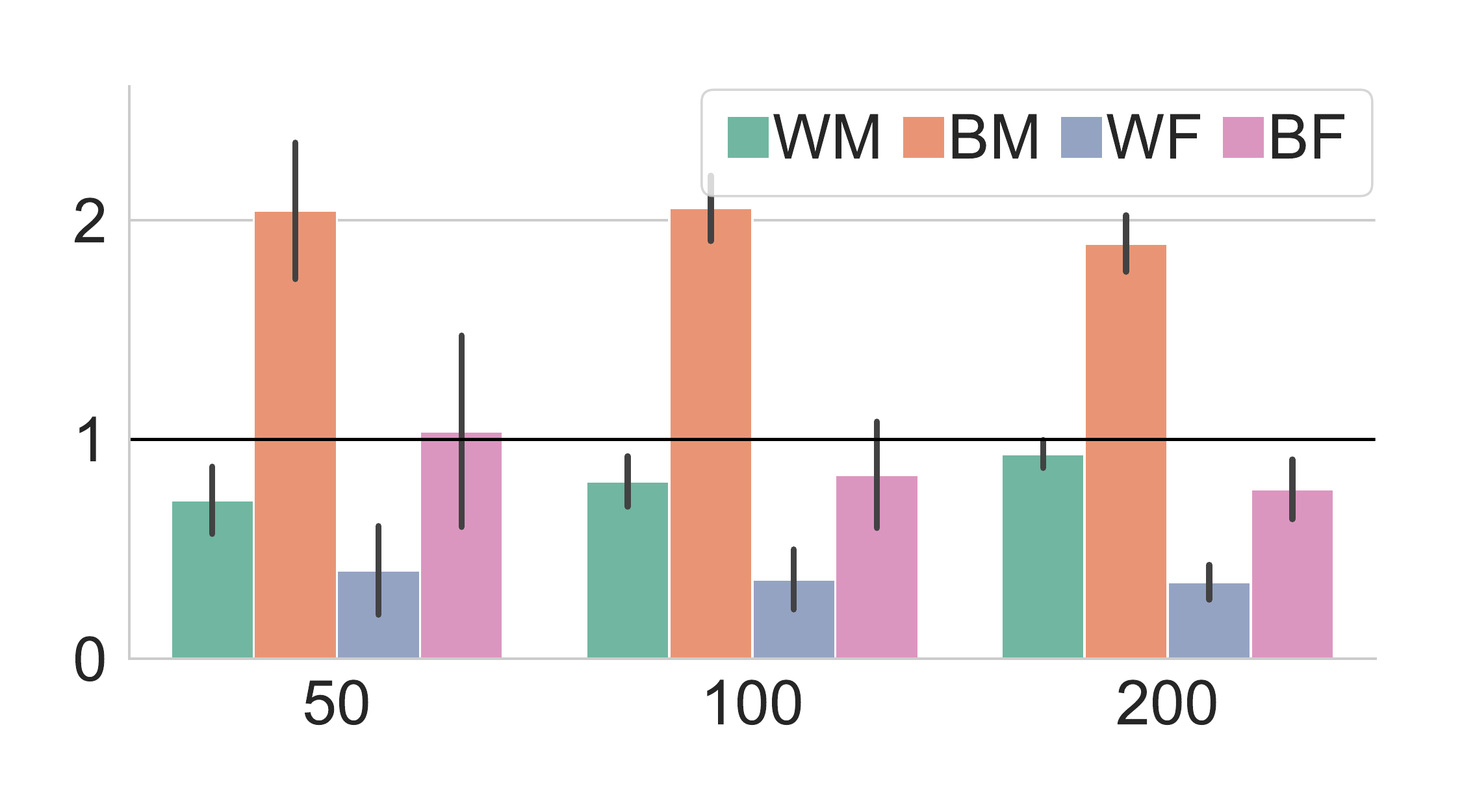}\label{fig:mvr_dp_Y_count_resolve}
	}
	\subfloat[quotas on $R$]{
	\includegraphics[width=0.33\linewidth]{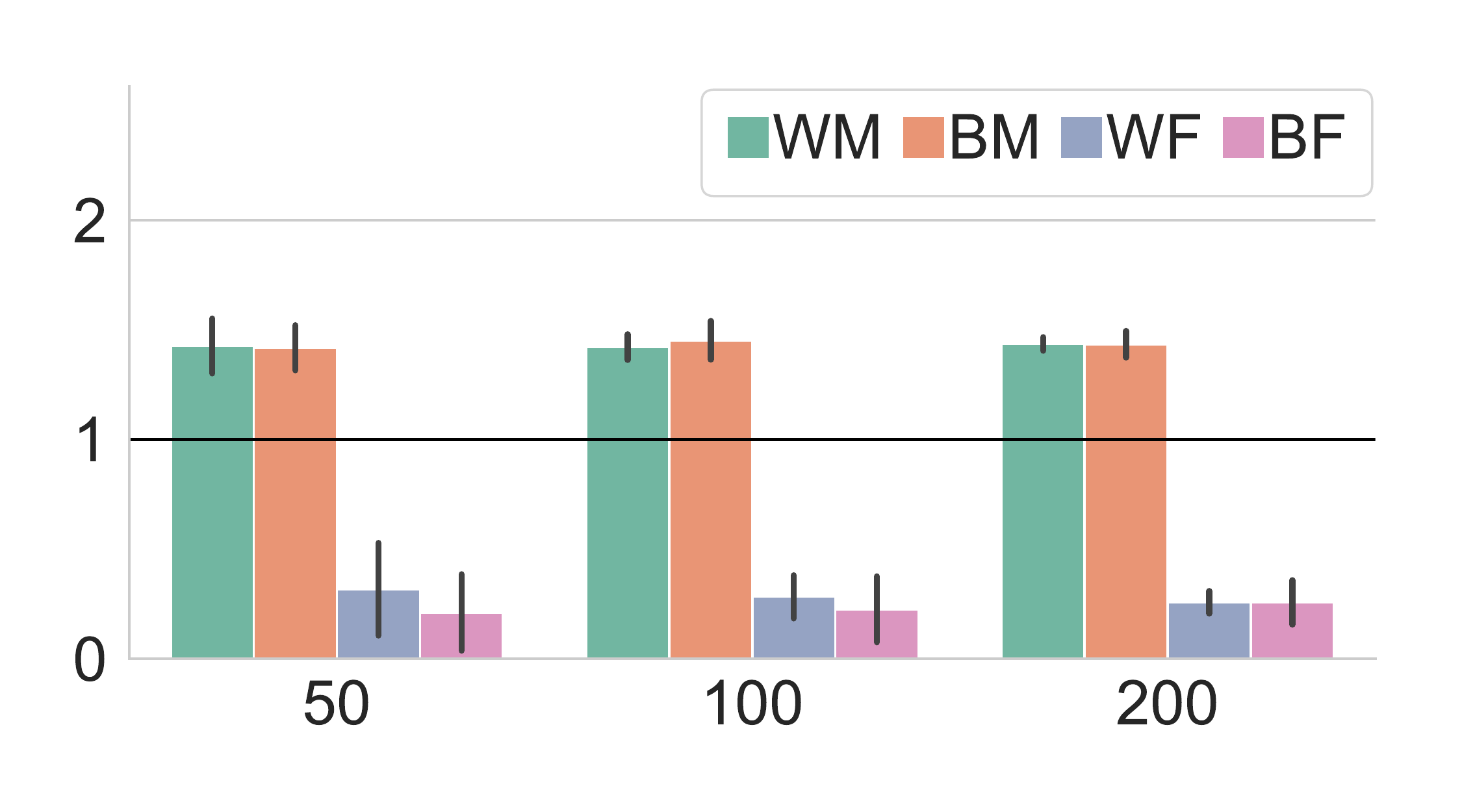}\label{fig:mvr_dp_Y_quotas_R}
	}
	\caption{Demographic parity on the \MVR dataset, generated using causal model $\mathcal{M}_1$ in Figure 2a with gender $G$, race $R$, resolving mediator weight-lifting ability $X$, and qualification $Y$.}
	\label{fig:mvr_dp_Y_all}
\end{figure*}

\begin{figure*}[h!]
	\centering
	\subfloat[original ranking]{\includegraphics[width=0.33\linewidth]{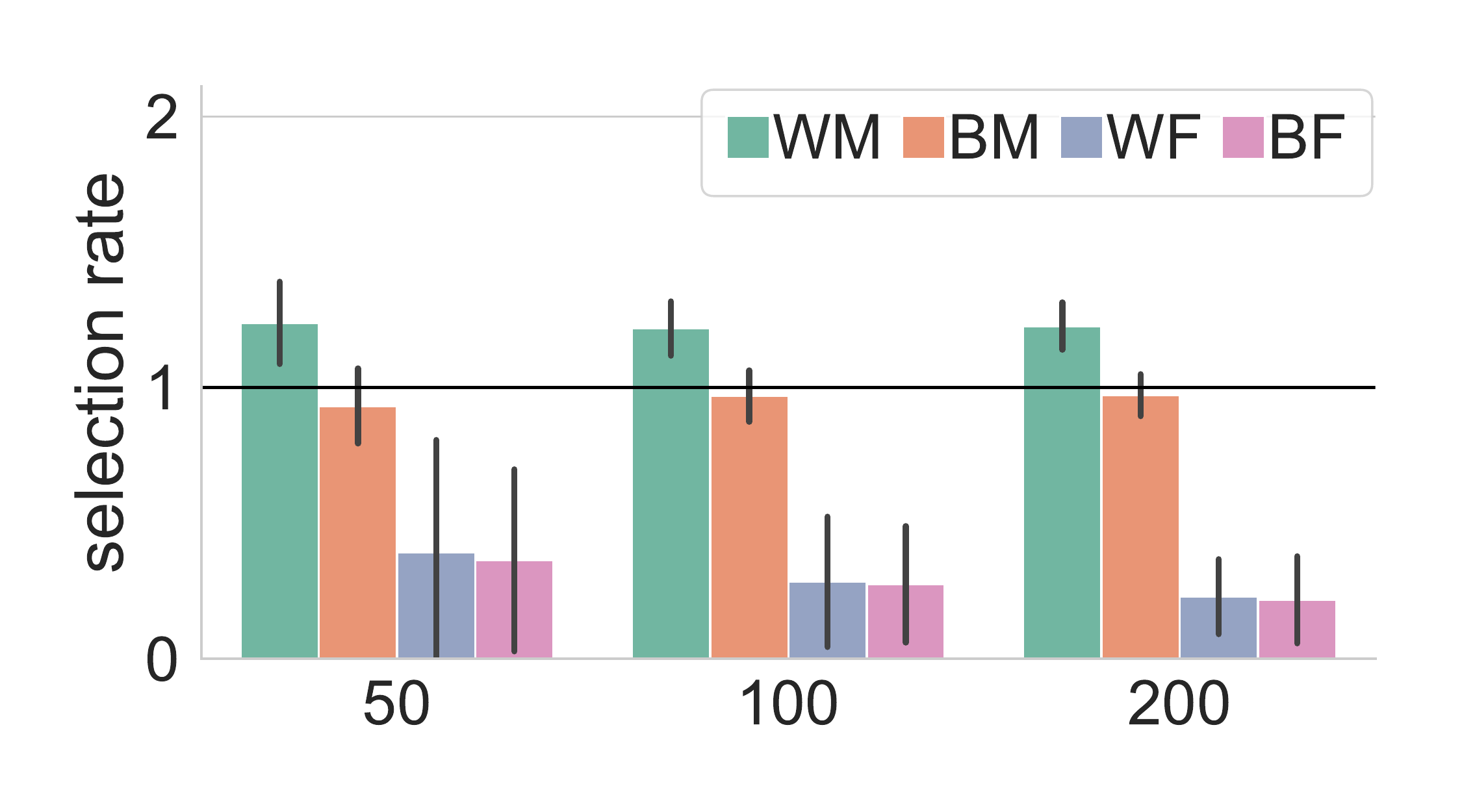}
	\label{fig:mvp_dp_Y}
	}
	\subfloat[\cfr]{
	\includegraphics[width=0.33\linewidth]{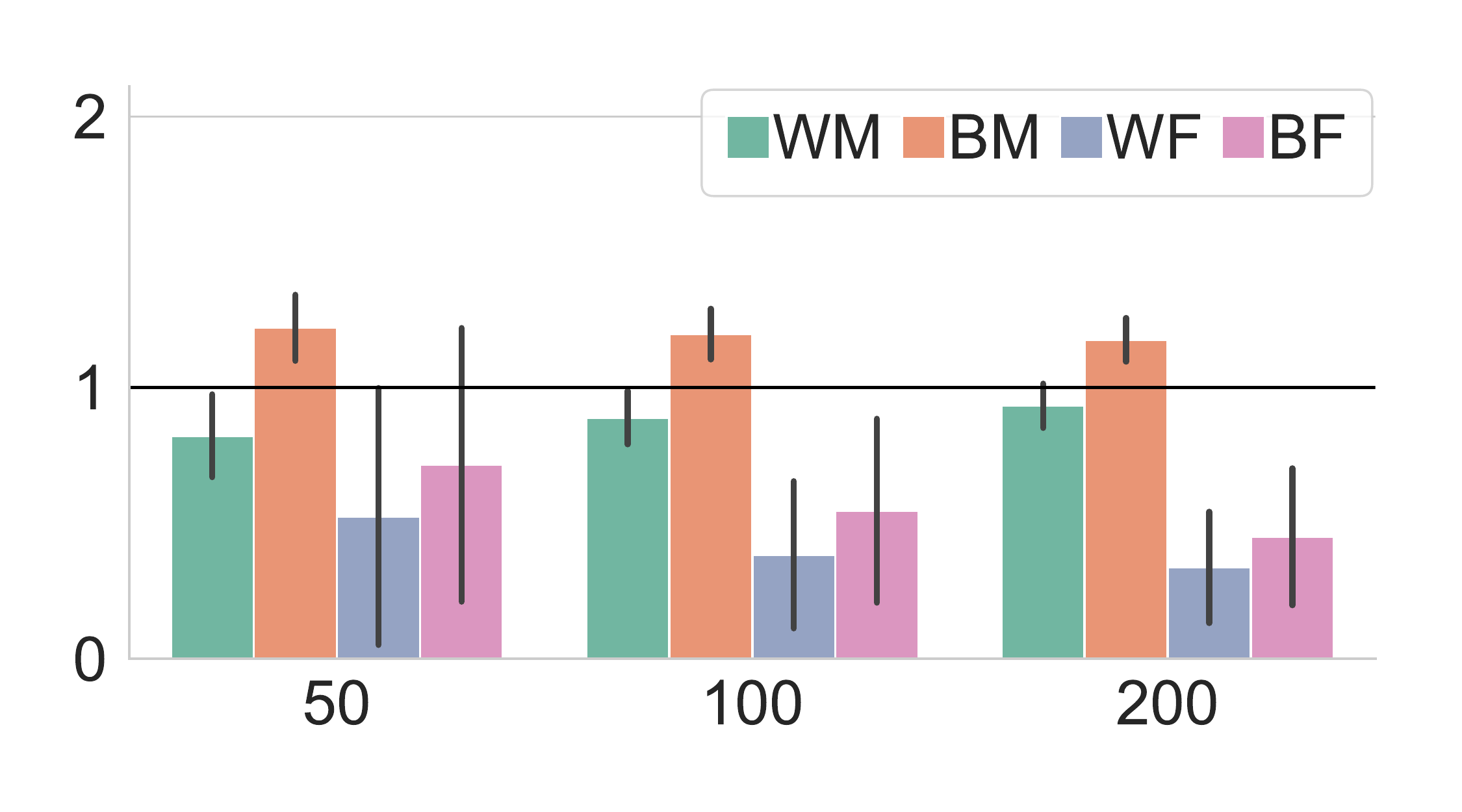}\label{fig:mvp_dp_Y_count_resolve}
	}
	\subfloat[quotas on $R$]{
	\includegraphics[width=0.33\linewidth]{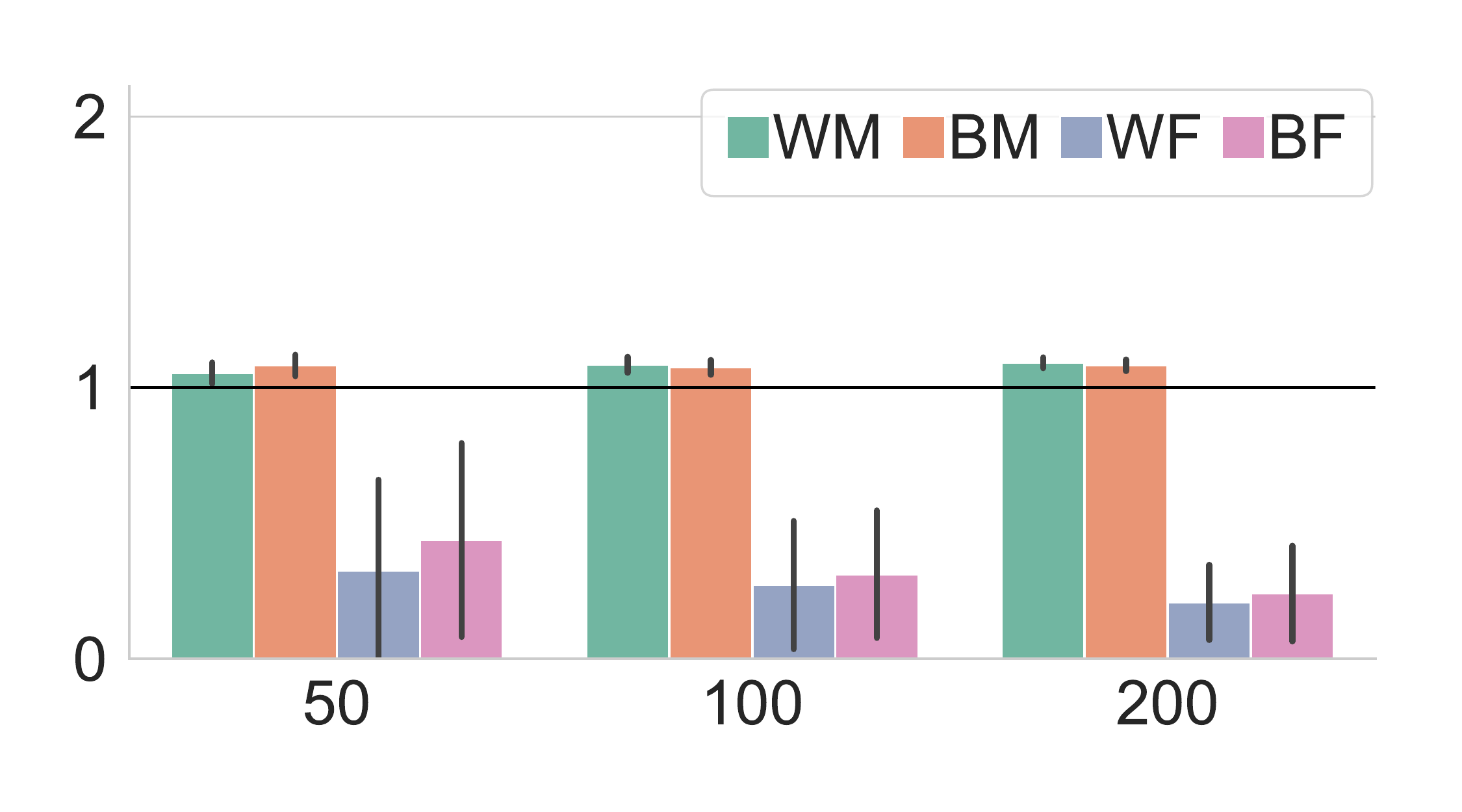}\label{fig:mvp_dp_Y_quotas_R}
	}
	\caption{Demographic parity on the \MVP dataset, generated using causal model $\mathcal{M}_1$ in Figure 2a with gender $G$, race $R$, resolving mediator weight-lifting ability $X$, qualification $Y$.}
	\label{fig:mvp_dp_Y_all}
\end{figure*}

\subsection{Score-based Ranking}
\label{sec:exp:score}


\paragraph{Moving company}

In the \MV example, the goal is to compute a ranking of the applicants on their qualification score $Y$ that is free of racial discrimination, while  allowing for a difference in weight-lifting ability $X$ between gender groups, thus treating $X$ as a resolving mediator.  

Figure~\ref{fig:mv_dp} shows selection rates of intersectional groups at the top-$k$ for $k=50,100,200$ for three treatments:  original ranking, counterfactually fair ranking that treats weight lifting ability $X$ as a resolving mediator, and a ranking that enforces representation constraints on race. Observe that the original ranking in Figure~\ref{fig:mv_dp_Y} under-represents women (WF and BF) compared to their proportion in the input, and that White men (WM) enjoy a higher selection rate than do Black men (BM).  Specifically, there are between 62-64\% \WM (40\% in the input), 27-28\% \BM (23\% in the input), 6\% \WF (23\% in the input), and 3-9\% \BF (14\% in the input) for the values of $k$ considered in this experiment. 

In comparison, as shown in Figure~\ref{fig:mv_dp_Y_count_resolve}, selection rates are higher for the Blacks, of both genders, than for the Whites.  For example, selection rate for White men is just over 1, while for Black men it's 1.5.  Selection rates also differ by gender, because weight-lifting ability $X$ is a mediator, and it encodes gender differences.  Finally, Figure~\ref{fig:mv_dp_Y_quotas_R} shows demographic party for intersectional groups when the ranking is computed using representation constraints (quotas) on race $R$: independently sorting  Black and White applicants on $Y$ and selecting the top individuals from each list in proportion to that group's representation in the input.  Opting for quotas on race rather than on gender, or on a combination of gender and race, is reasonable here, and it implicitly encodes a normative judgement that is explicit in our causal model: that race should not impact the outcome, while gender may.   

Figure~\ref{fig:mvr_dp_Y_all} shows demographic parity for \MVR.  We observe that selection rates are higher for the Blacks of both genders in the counterfactually fair ranking in Figure~\ref{fig:mvr_dp_Y_count_resolve} than in the original ranking (Figure~\ref{fig:mvr_dp_Y}) or in the ranking with repreentation constraints on race (Figure~\ref{fig:mvr_dp_Y_quotas_R}). Further, comparing Figure~\ref{fig:mvr_dp_Y_count_resolve} with Figure~\ref{fig:mv_dp_Y_count_resolve}, we see that selection rates for \BM are increased more significantly in \MVR than in \MV,  as expected.  

Figure~\ref{fig:mvp_dp_Y_all} shows demographic parity for \MVP.  Recall that this dataset has a higher proportion of male applicants in the input, and that \BM constitute the largest intersectional group, 50\%.  In this dataset the original ranking (Figure~\ref{fig:mvp_dp_Y}) is still favoring \WM over \BM, but not as strongly as in \MV and \MVR.  The counterfactually fair ranking (Figure~\ref{fig:mvp_dp_Y_count_resolve}) places Black applicants at the top-$k$ in higher proportion, but the magnitude of the difference is smaller than for \MV and \MVR, as expected. 

\begin{figure*}[t!]
	\centering
	\subfloat[\MV]{\includegraphics[width=0.33\linewidth]{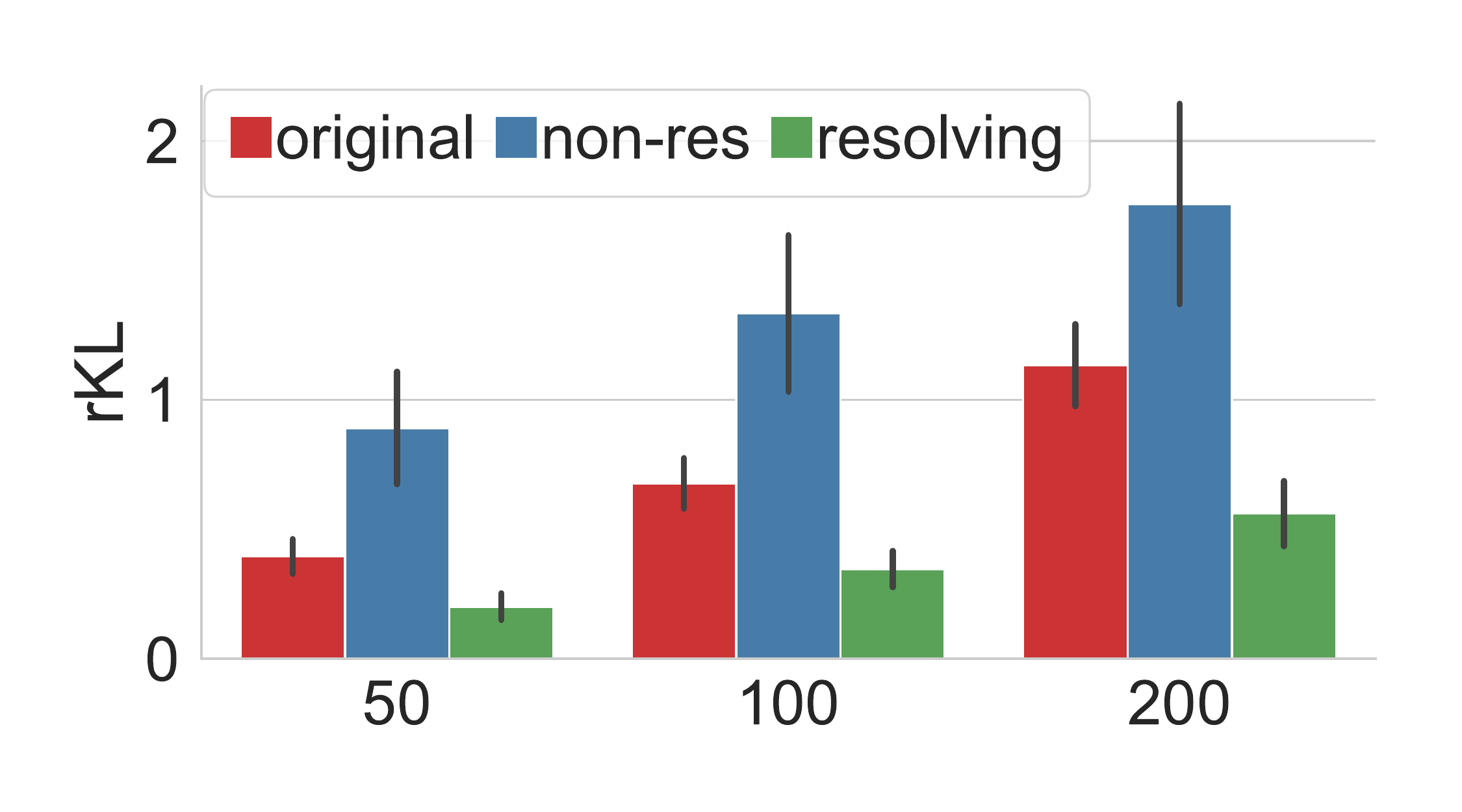}
	\label{fig:mv_dp_rKL}
	}
	\subfloat[\MVR]{
	\includegraphics[width=0.33\linewidth]{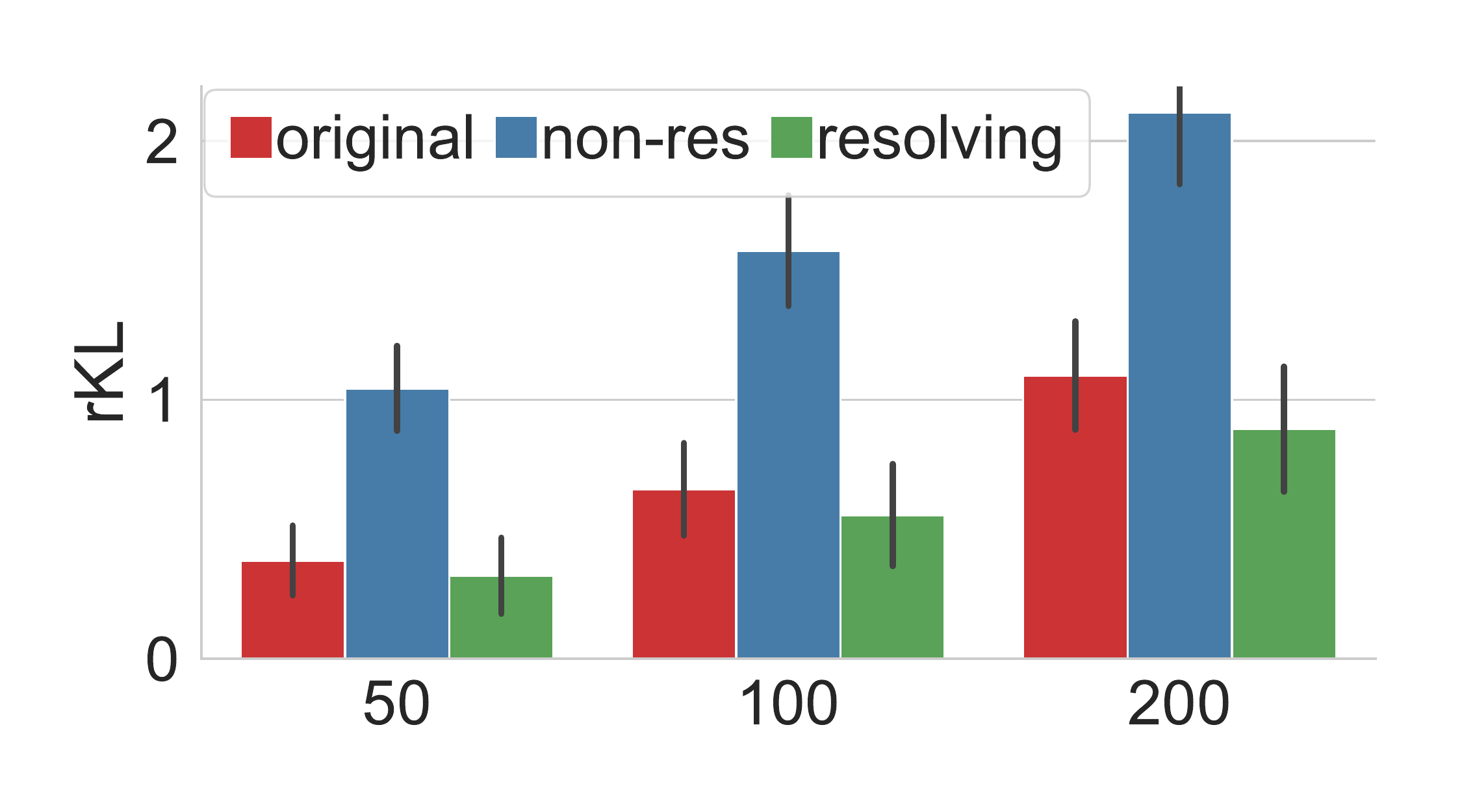}\label{fig:mvr_dp_rKL}
	}
	\subfloat[\MVP]{
	\includegraphics[width=0.33\linewidth]{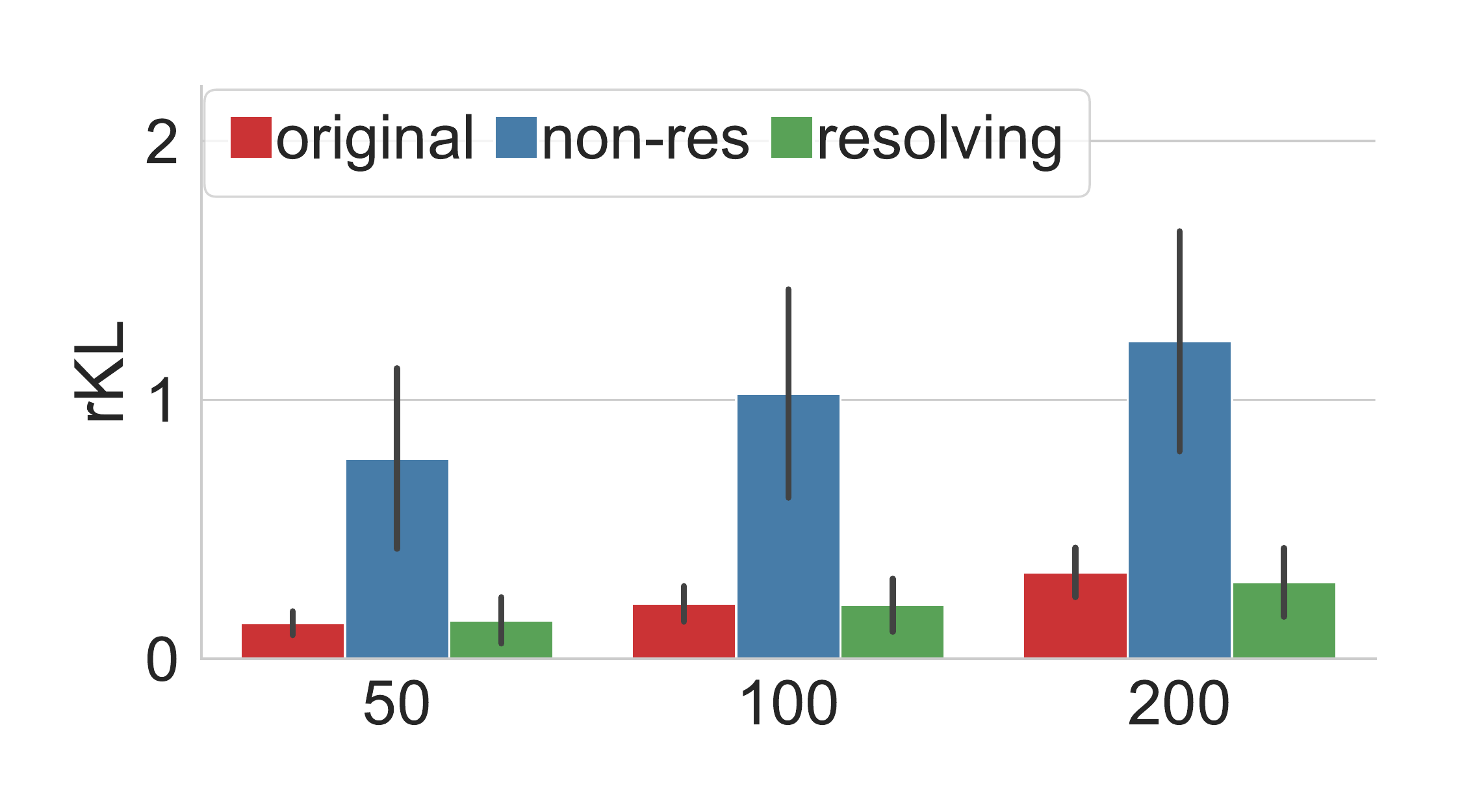}\label{fig:mvp_dp_rKL}
	}
	\caption{\rKL on \MV, \MVR, and \MVP datasets.  Lower values of \rKL show that selection rates match proportions in the population at every prefix of the ranking.}
	\label{fig:mv_rKL}
\end{figure*}

Figure~\ref{fig:mv_rKL} shows normalized discounted KL-divergence (\rKL) on all  \MV datasets.  Recall that \rKL is a fairness measure that compares selection rates of intersectional groups to their proportion in the input; this measure computes these comparisons progressively through prefixes of a ranked list.  A lower value of \rKL (0 or close to 0) signifies that every intersectional group is selected in proportion to their representation in the input, in every prefix of the ranked list.  Observe from Figure~\ref{fig:mv_rKL} that \rKL is lower for the counterfactually fair rankings that treats weight-lifting ability $X$ as a resolving mediator than for the corresponding original rankings.   The value of \rKL is higher when $X$ is treated as non-resolving.  We consider treating $X$ as resolving more reasonable in this scenario, since weight-lifting ability should impact whether an applicant is hired for by a moving company.  

We also computed $Y$-utility loss at top-$k$, based on the original $Y$ scores (see Section~\ref{sec:exp:methods} for details). We found that counterfactually fair rankings for \MV suffer 3-4\% loss across the values of $k$, slightly higher than the loss of rankings with quotas on race, which is at most 1\%.  $Y$-utility loss on \MVR and \MVP is comparable to that on \MV, ranging between 2\% and 10\%,  and between  1\% and 5\%, respectively.

\paragraph{\CM}

\begin{figure*}[h]
	\centering
	\subfloat[original ranking]{\includegraphics[width=0.33\linewidth]{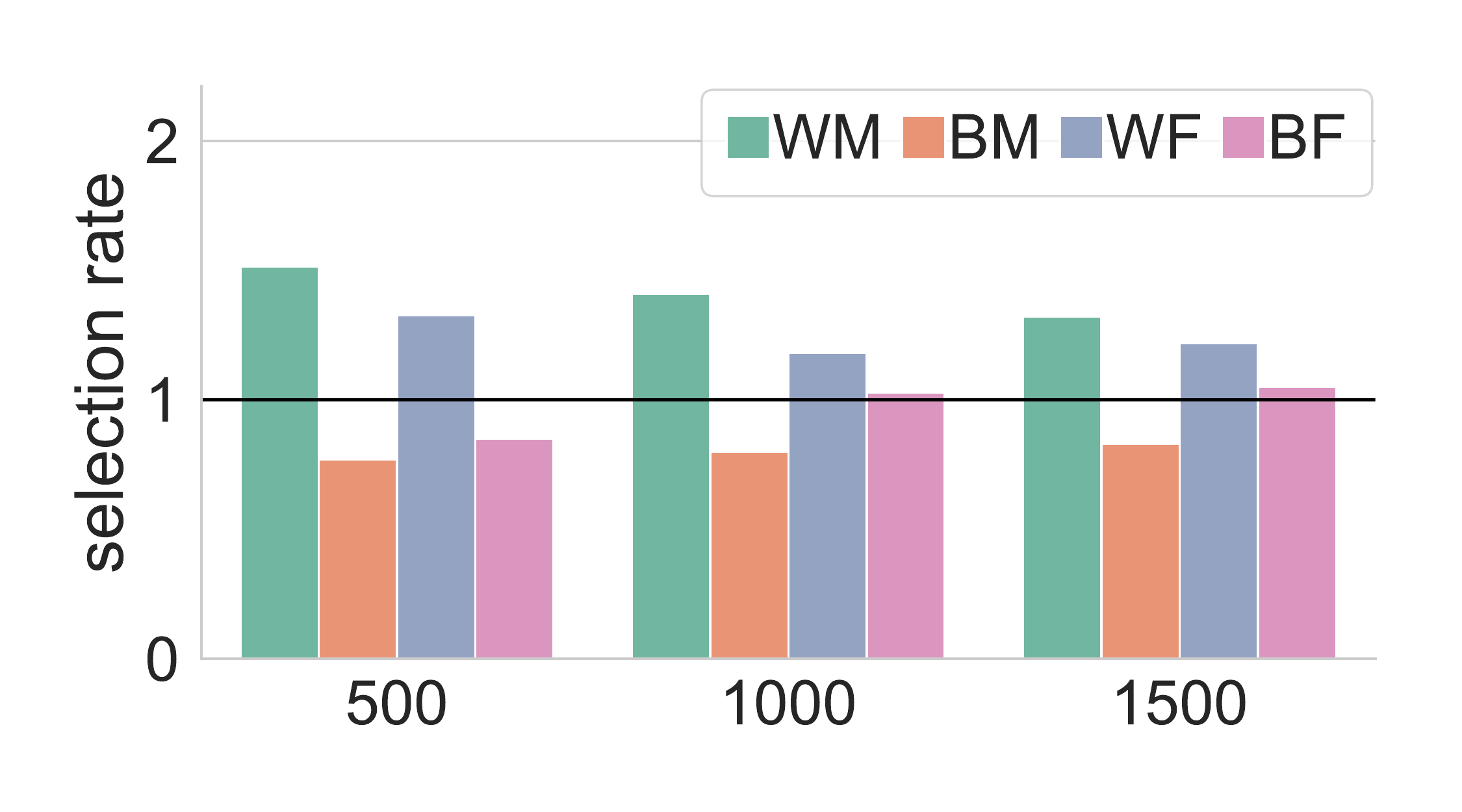}\label{fig:cm_dp_Y}
	}
	\subfloat[resolving $X$]{
	\includegraphics[width=0.33\linewidth]{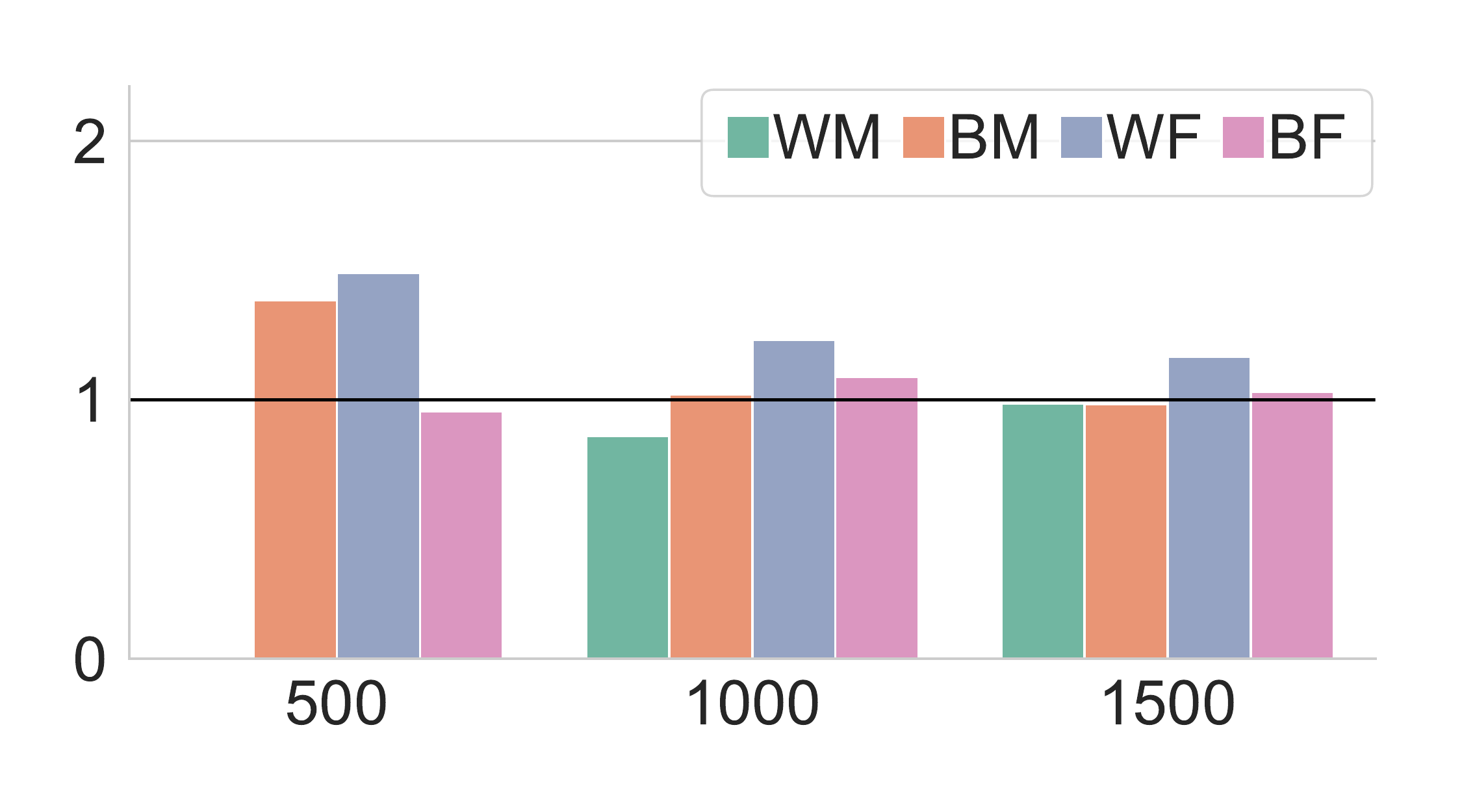}\label{fig:cm_dp_Y_count_resolve}
	}
	\subfloat[non-resolving $X$]{
	\includegraphics[width=0.33\linewidth]{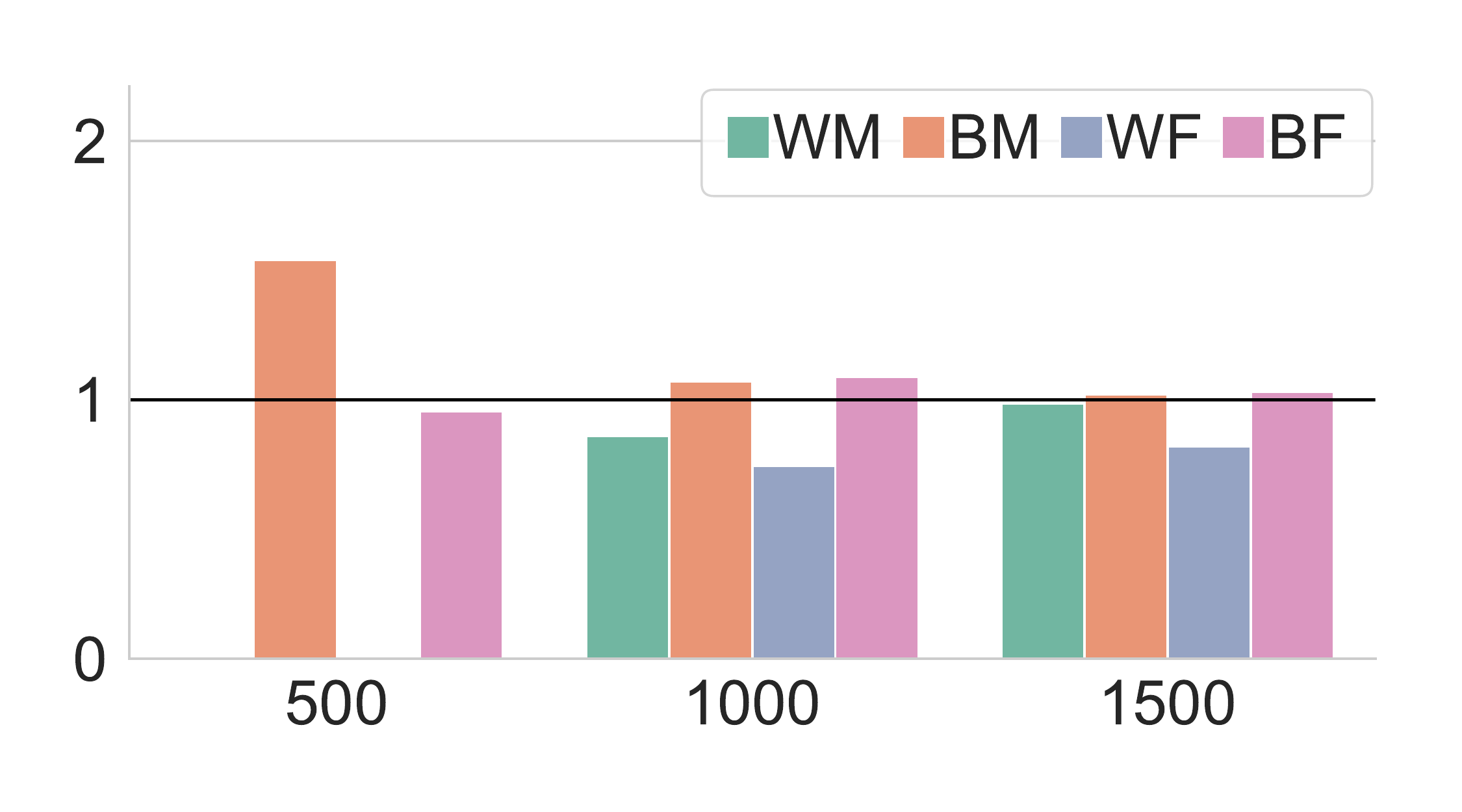}\label{fig:cm_dp_Y_count}
	}
	\caption{Demographic parity on \CM, with causal model  $\mathcal{M}_1$  in Figure~\ref{fig:cm_covariate}, with gender $G$, race $R$, prior arrests $X$, and decile score $Y$.  Figures (b) and (c) present counterfactually fair rankings, differing in whether $X$ is treated as resolving.}
	\label{fig:cm_dp}
\end{figure*}

\begin{figure*}[t]
	\centering
	\subfloat[\CM]{
	\includegraphics[width=0.33\linewidth]{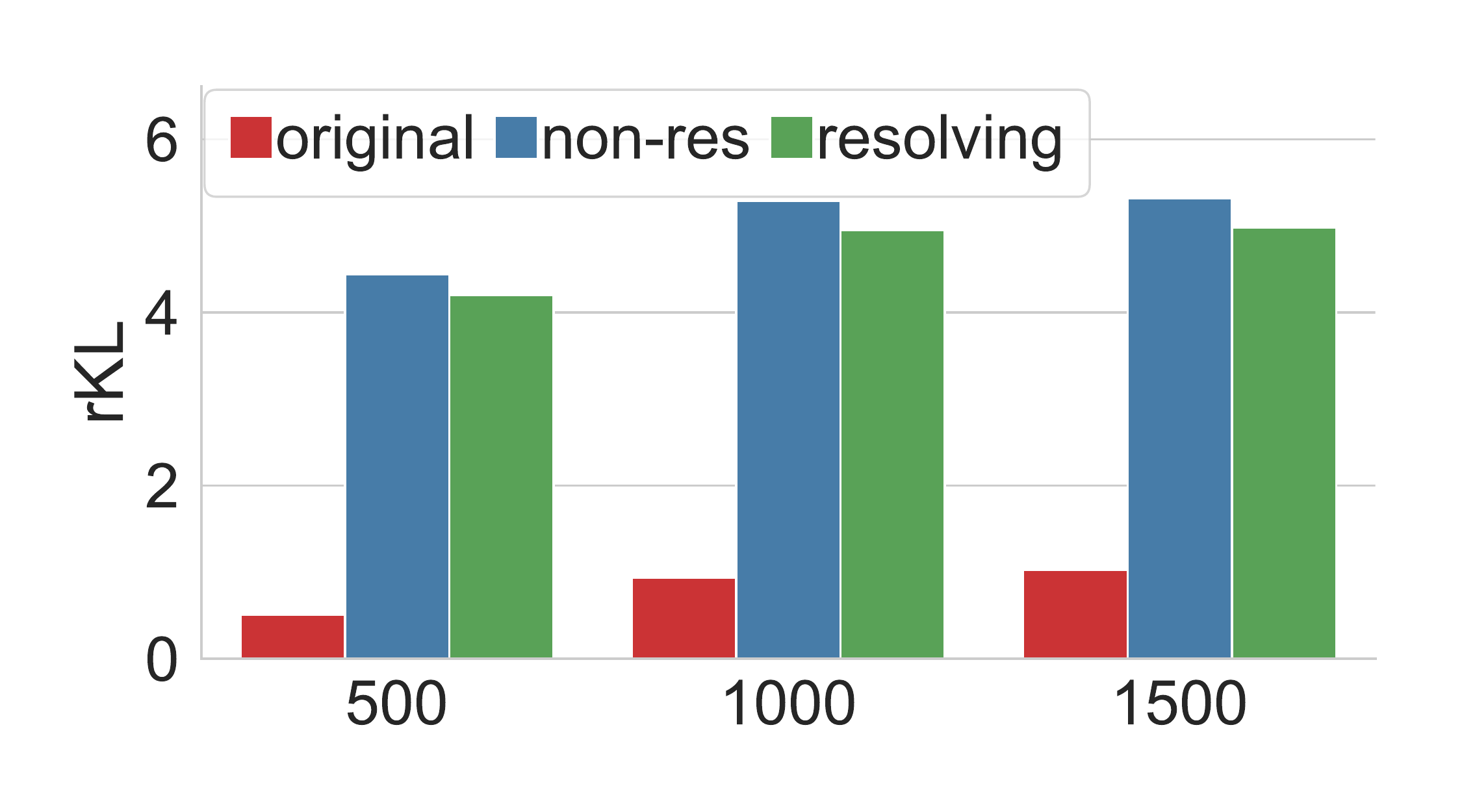}\label{fig:cm_dp_rKL}
	}
	\subfloat[\MP]{\includegraphics[width=0.33\linewidth]{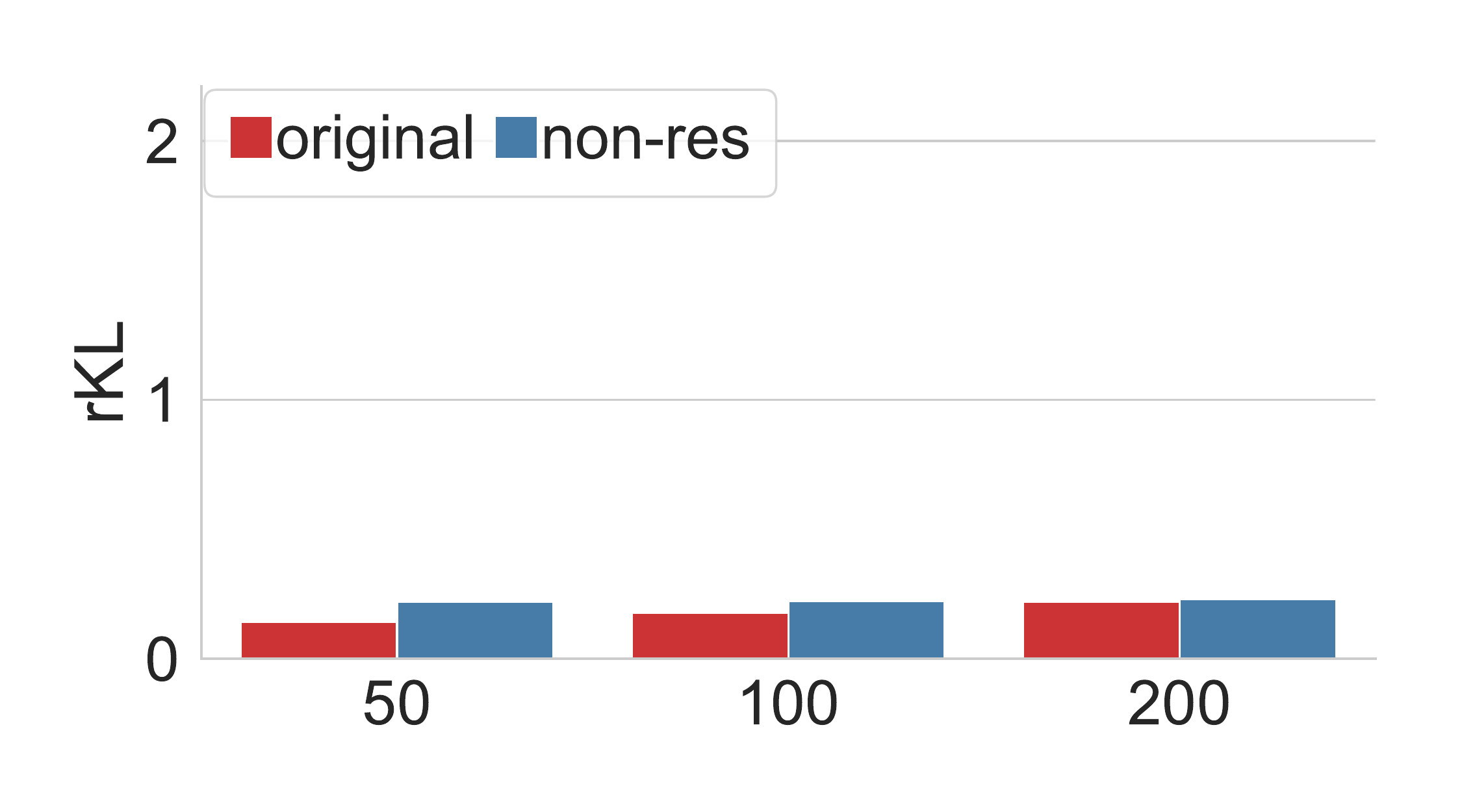}
	\label{fig:mp_dp_rKL}
	}
	\caption{\rKL on the real datasets \MP and \CM.  Lower values of \rKL show that selection rates match proportions in the population at every prefix of the ranking.  Red bars represent the original ranking, blue bars represent counterfactually fair rankings that treat weight lifting ability $X$ as a non-resolving mediator, gree bars represent counterfactually fair rankings that treat weight lifting ability $X$ as a resolving mediator.}
	\label{fig:real_rKL}
\end{figure*}

In \CM we rank defendants on their scores from lower to higher, prioritizing them for release, or for access to supportive services, as part of a comprehensive reform of the criminal justice system.  Our goal is to produce a ranking that is free of racial and gender discrimination.  In Figure~\ref{fig:cm_dp_Y} we show selection rates of intersectional groups by race and gender in the original ranking, and observe that Whites of both genders are selected at much higher rates than Blacks.  Gender has different effect by race: males are selected at higher rates for Whites, and at lower rates for Blacks.  There are 33-38\%  White males (25\% in the input), 46-49\% Black males (59\% in the input), 7-8\% White females (6\% in the input), and 8-10\% Black females (10\% in the input),  for $k= 500, 1000, 1500$.

There is some debate about whether the number of prior arrests, $X$, should be treated as a resolving variable. We compare demographic parity for intersectional groups for two options: Figures~\ref{fig:cm_dp_Y_count_resolve} treats $X$ as a resolving variable, while~\ref{fig:cm_dp_Y_count} does not.  By treating $X$ as non-resolving, we are stating that the number of prior arrests is itself subject to racial discrimination.  Comparing Figures~\ref{fig:cm_dp_Y_count_resolve} to Figure~\ref{fig:cm_dp_Y}, we observe an increase in selection rates for Black males and Black females, and a significant reduction in selection rates for White males.  Further, comparing with Figure~\ref{fig:mv_dp_Y_quotas_R}, we observe that only Black individuals are present at the top-$500$, and that selection rates for larger values of $k$ are close to 1, achieving demographic parity.

Figure~\ref{fig:cm_dp_rKL} presents \rKL on \CM.  The value  is higher for both counerfactually fair treatments (number of prior arrests $X$ treated as resolving and non-resolving, respectively).  This is as expected, since \BM are placed at the top-$k$ at higher rates by the counterfactually fair methods than is warranted by their proportion in the input alone.

We also computed utility loss at top-$k$, based on the original $Y$ scores, and found that overall utility loss is low in most cases, ranging between 3\% and 8\% when $X$ is treated as a resolving variable, and between 3\% and 10\% when $X$ is treated as non-resolving.  The slightly higher loss for the latter case is expected, because we are allowing the model to correct for historical discrimination in the data more strongly in this case, thus departing from the original ranking further.  As for the \MV example, we found no loss in in-group utility, because our method does not reorder entities within an intersectional group.

\paragraph{\ratio} Table~\ref{tab:exp_ratio} shows in-group fairness ratio (\ratio) on all \MV datasets and on \CM. For \CM, we broke ties by $Y$-score by randomly permuting the items within an equivalence class.  Recall that a higher value of \ratio is better, since it indicates that the ratio of scores of the lowest-scoring selected item among the top-$k$ and of the highest-scoring item not among the top-$k$ is close to 1.  Observe that most \ratio values are close to 1, meaning that there is only a limited amount of re-ordering of individuals within each intersectional group. Further, \igf loss is balanced among intersetional groups in all except two cases, both for \MVP.  This warrants further investigation of this  type of an intersectional effect.
\begin{table*}[h!]
	\centering
	\resizebox{0.98\linewidth}{!}{
		\begin{tabular}{||c|c|c|c|c|c|c|c|c|c|c|c|c|c||}
		    \cline{1-14}
			\multirow{2}{*}{Dataset} & \multirow{2}{*}{Ranking} & \multicolumn{4}{|c|}{$k_1$} & \multicolumn{4}{|c|}{$k_2$} & \multicolumn{4}{|c|}{$k_3$} \\
			\cline{3-14}
			&  &  WM & BM & WF & BF &  WM & BM & WF & BF &  WM & BM & WF & BF \\
			\cline{1-14}
			
			\multirow{2}{*}{\MV} &    non-res &   0.99 &  0.99 &  0.92 &  0.94 &   0.98 &  0.96 &  0.93 &  0.92  &  0.96 &  0.96 &  0.89 &  0.88 \\
			\cline{2-14}
			&  resolving &   0.96 &  0.95 &  0.99 &  0.99 &  0.95 &  0.94 &  0.97 &  0.96 &   0.93 &  0.92 &  0.92 &  0.95 \\
			\cline{0-13}

			\multirow{2}{*}{\MVR} &    non-res &   0.99 &  0.98 &  0.92 &  0.93 &   0.97 &  0.95 &  0.90 &  0.90 &   0.93 &  0.92 &  0.88 &  0.89 \\
			\cline{2-14}
			&  resolving &   0.96 &  0.94 &  0.98 &  0.97 &    0.93 &  0.91 &  0.95 &  0.93 &  0.92 &  0.90 &  0.93 &  0.93 \\
			\cline{0-13}

			\multirow{2}{*}{\MVP} &    non-res &  0.98 &  0.95 &  0.99 &  0.96 &   0.96 &  0.94 &  0.96 &  0.94 &     0.94 &  0.91 &  0.96 &  0.92 \\
			\cline{2-14}
			&  resolving & {\bf 0.97} &  {\bf 0.94} &  {\bf 0.50} &  {\bf 0.59} &  {\bf 0.96} &  {\bf 0.93} &  {\bf 0.80}
			&  {\bf 0.80} &  0.94 &  0.91 &  0.89 &  0.88 \\
			\cline{0-13}

			\multirow{2}{*}{\CM} &    non-res &   N/A &  1.00 & N/A &  1.00 &   1.00 &  1.00 &  1.00 &  1.00 &   1.00 &  1.00 &  1.00 &  1.00 \\
			\cline{2-14}
			&  resolving &   N/A &  1.00 &  1.00 &  1.00 &   1.00 &  1.00 &  1.00 &  1.00 &   1.00 &  1.00 &  1.00 &  1.00 \\
			\cline{0-13}

			\MP & non-res &  {\bf 0.38} &   {\bf 0.22} &  {\bf  0.2} &   {\bf 0.74} &  0.32 &  0.22 &  0.2 &  0.21 &  0.32 &  0.22 &  0.2 &  0.21 \\
			\cline{0-13}

	\end{tabular}}
	\caption{\ratio on \MV variants and on \CM and \MP. A higher value is better: it indicates that the ratio of scores of the lowest-scoring selected item among the top-$k$ and of the highest-scoring item not among the top-$k$ is close to 1.  In the table, $k_{1,2,3}=50,100,200$ for \MV, \MVR, \MVP ($n=2000$), and \MP ($n=15,675$) and $k_{1,2,3}=500,1000,1500$ for \CM ($n=4162$).  N/A is used when a particular intersectional group is not represented among the top-$k$.  Treatments where an imbalance in \ratio is observed among intersectional groups are highlighted in {\bf bold font}.}
	\label{tab:exp_ratio}
\end{table*}

\begin{figure*}[t!]
	\centering
	\subfloat[original ranking]{\includegraphics[width=0.33\linewidth]{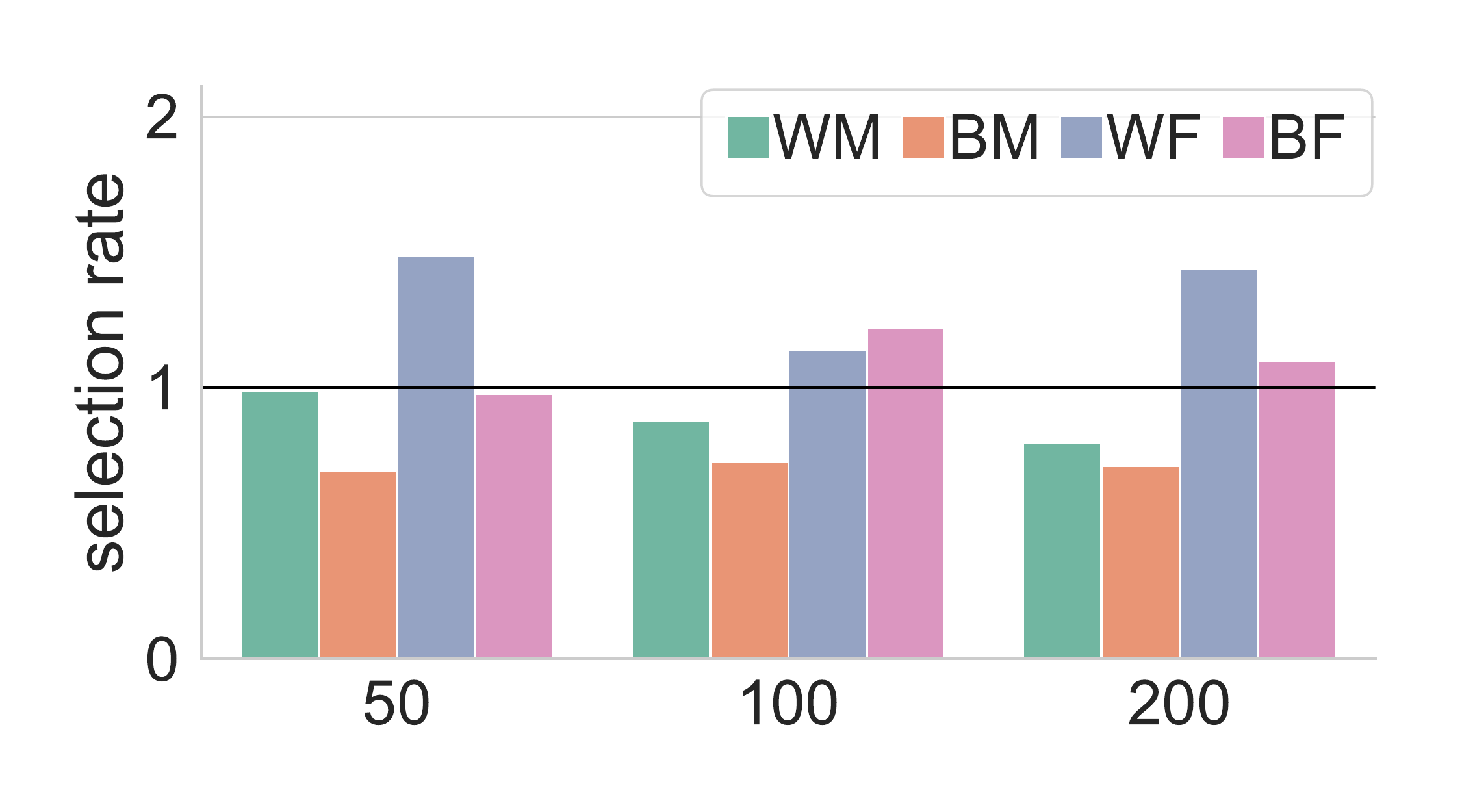}
	\label{fig:mp_dp_Y}
	}
	\subfloat[\cfr]{
	\includegraphics[width=0.33\linewidth]{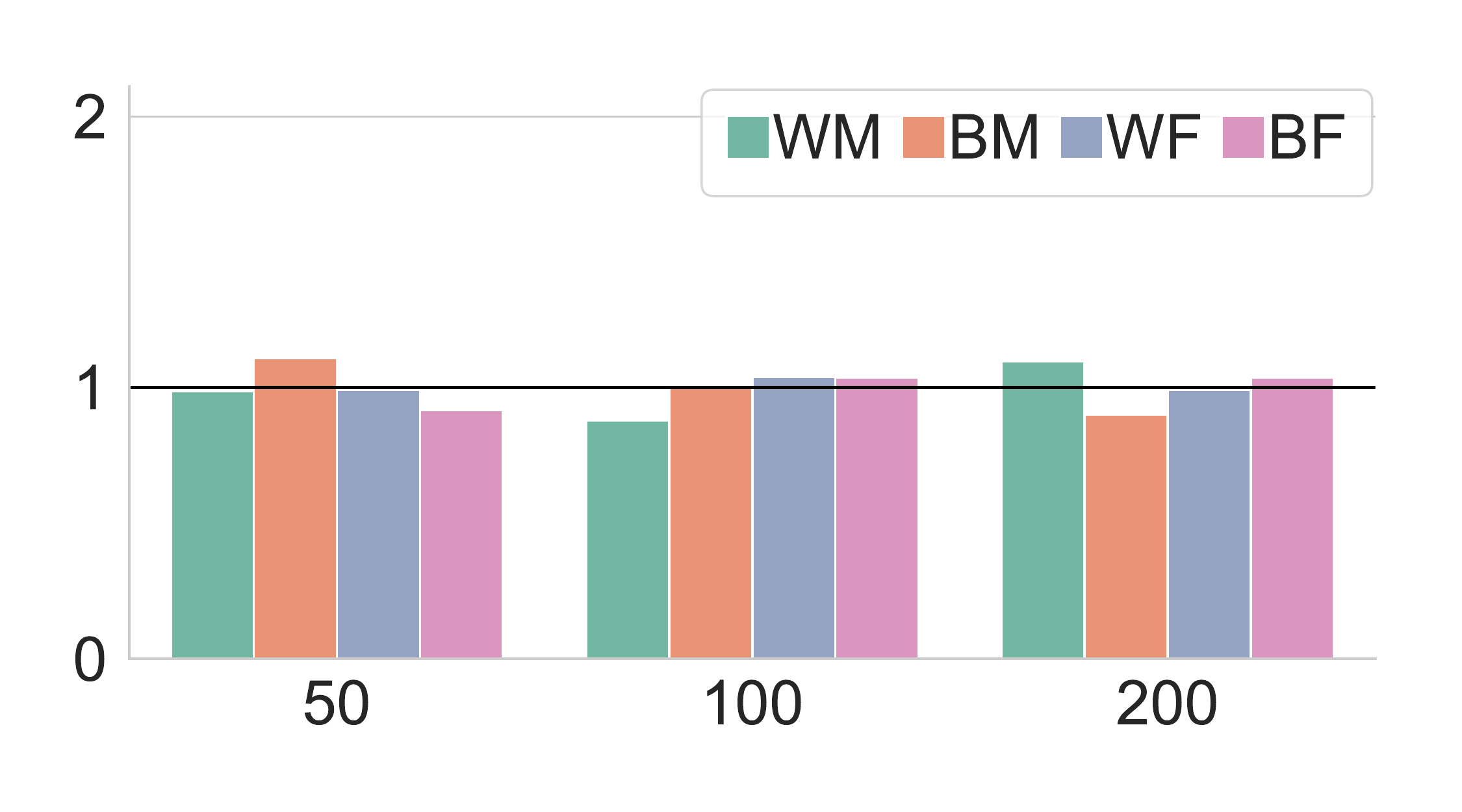}
	\label{fig:mp_dp_Y_count}
	}
	\subfloat[quotas on $R$]{
	\includegraphics[width=0.33\linewidth]{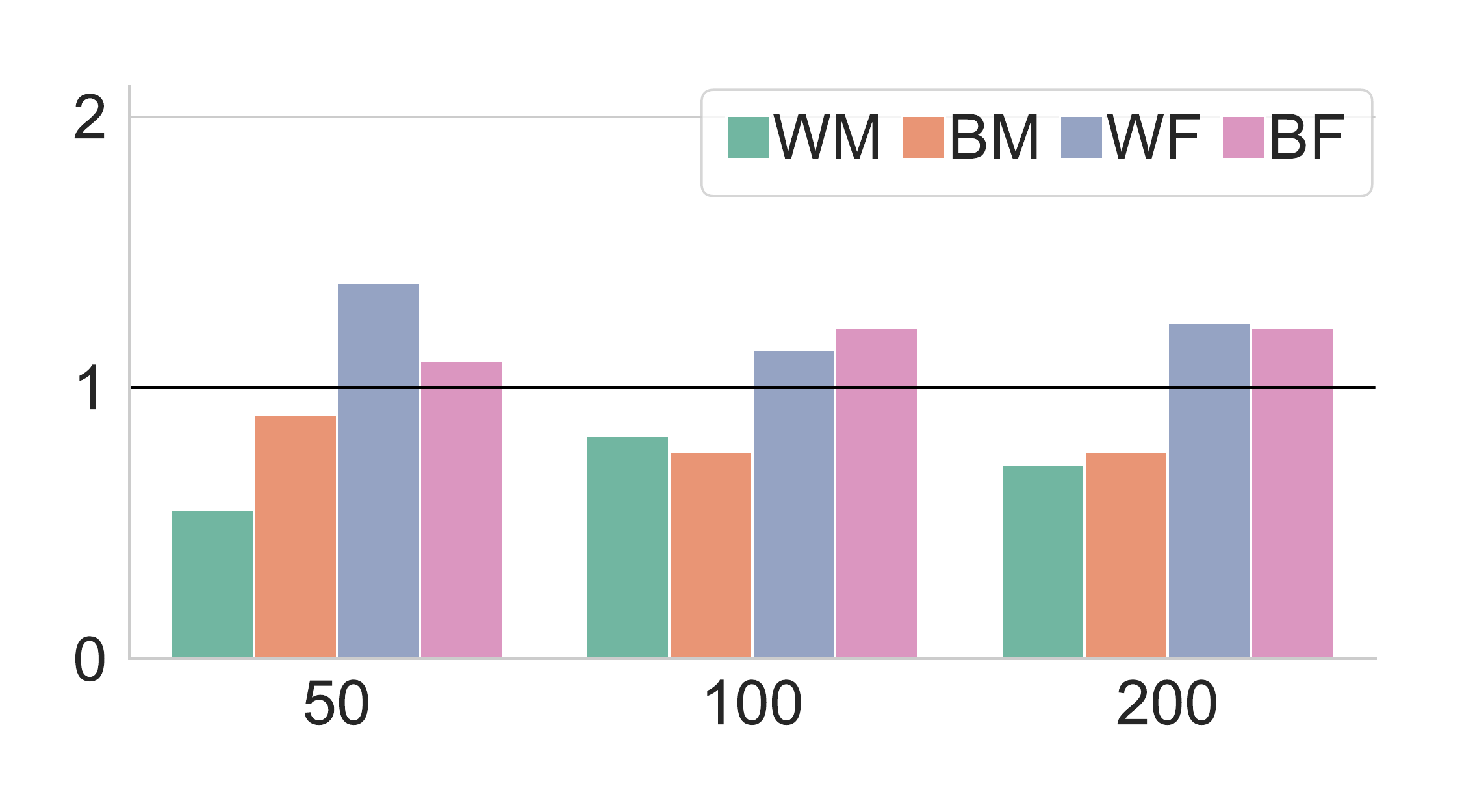}
	\label{fig:mp_dp_Y_quotas_R}
	}
	\caption{Demographic parity on the \MP dataset, generated using causal model $\mathcal{M}_1$ in Figure 2a, with mediator $X$ representing age that is a moderator. In $\mathcal{M}_1$, gender $G$, race $R$, and age $X$ causally affect healthcare utilization score $Y$. Ranking is on $Y$ from higher to lower, $k$ takes on values 50, 100, and 200.}
	\label{fig:mp_dp_Y_all}
\end{figure*}

\paragraph{\MP} Figure~\ref{fig:mp_dp_Y_all} shows selection rates of four intersectional groups on the \MP dataset.  Recall that the causal model for \MP is different from other models we considered in this paper in two ways.  First, $X$ is a moderator rather than a mediator, and, second, healthcare utilization $Y = f(X, G, R)$ is a non-linear function of age $X$, allowing different nonlinear functions $f_{g,r}(X)$ for each intersectional group. Consider Figure~\ref{fig:mp_dp_Y} and observe that selection rates are higher for Whites than for Blacks, and that they are highest for \WF and lowest for \BF. The counterfactually fair ranking in Figure~\ref{fig:mp_dp_Y_count} balances the selection rates across groups, substantially increasing them for \BF.  We also show results for a ranking that uses proportional representation (quotas) on race in Figure~\ref{fig:mp_dp_Y_quotas_R}, which is appropriate here because the original ranking shows a strong disparity in selection rates specifically by race.   Deciding which of these results is (more) reasonable requires domain knowledge, for example about the reasons and mechanisms causing different utilization, and we believe a transparent causal model can help organize that inquiry.

Another interesting comparison between the original and the counterfactually fair ranking on \MP is the average age at top-$k$ per intersectional group, shown in Table~\ref{tab:mp_age}.  Observe that average age varies among intersectional groups  more substantially in the original ranking than in the counterfactual.  Results like this may help identify specific cross-sections of intersectional groups experiencing more or less fair utilization rates.

Figure~\ref{fig:mp_dp_rKL} presents \rKL on \MP.  The value is low for both original and non-resolving (for age $X$) treatments for \MP.  

We also computed $Y$-utility loss for \MP and found that it ranges between 13.6\% and 18.8\% for $k=50,100,200$.  This is higher than for \CM (at most 4\%) and \MV 
(at most 10\%), and is due to the difference in the underlying score distributions.  Specifically, $Y$-scores range between 0 and 500 in \MP, dropping sharply after the top-20.  For COMPAS, $Y$-scores are in the $[0,10]$ range, and for \MV they are in the range $[-3,3]$.

\begin{table*}
	\centering
	\resizebox{0.98\linewidth}{!}{
		\begin{tabular}{||c|c|c|c|c|c|c|c|c|c|c|c|c||}
		    \cline{1-13}
			\multirow{2}{*}{Ranking} & \multicolumn{4}{c|}{top-$k_1$} & \multicolumn{4}{c|}{top-$k_2$} & \multicolumn{4}{c|}{top-$k_3$} \\
			\cline{2-13}
			 & WM & BM & WF & BF & WM & BM & WF & BF & WM & BM & WF & BF \\
			\cline{1-13}
			 original &  57.33 &  45.7 &  71.47 &  67.0 &  54.31 &  49.38 &  71.87 &  62.60 &  59.28 &  53.54 &  63.76 &  61.94 \\
			\cline{0-12}
			counter &  69.56 &  54.5 &  66.20 &  60.6 &  66.31 &  54.90 &  65.29 &  60.88 &  59.40 &  50.83 &  63.78 &  56.38 \\
			\cline{0-12}

	\end{tabular}}
	\caption{Average age on \MP, in which $k_{1,2,3}=50,100,200$ with $n=15,675$.}
	\label{tab:mp_age}
\end{table*}

\subsection{Learning to Rank}
\label{sec:exp:ltr}

\begin{figure*}[t!]
	\centering
	\subfloat[\MV, $k=500$]{\includegraphics[width=0.49\linewidth]{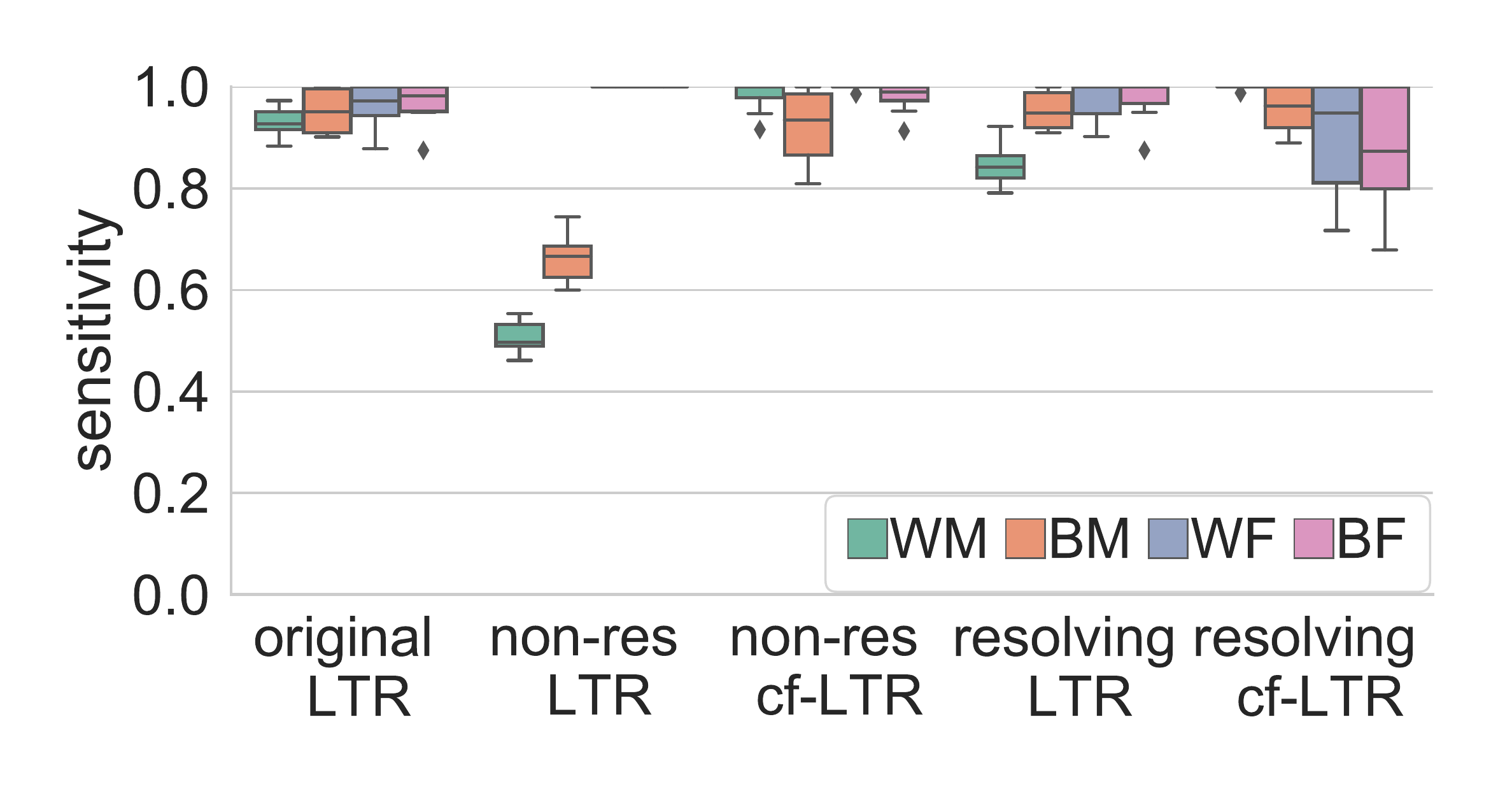}
	\label{fig:mv_eo_ltr}
	}
	\subfloat[\CM, $k=500$]{
	\includegraphics[width=0.49\linewidth]{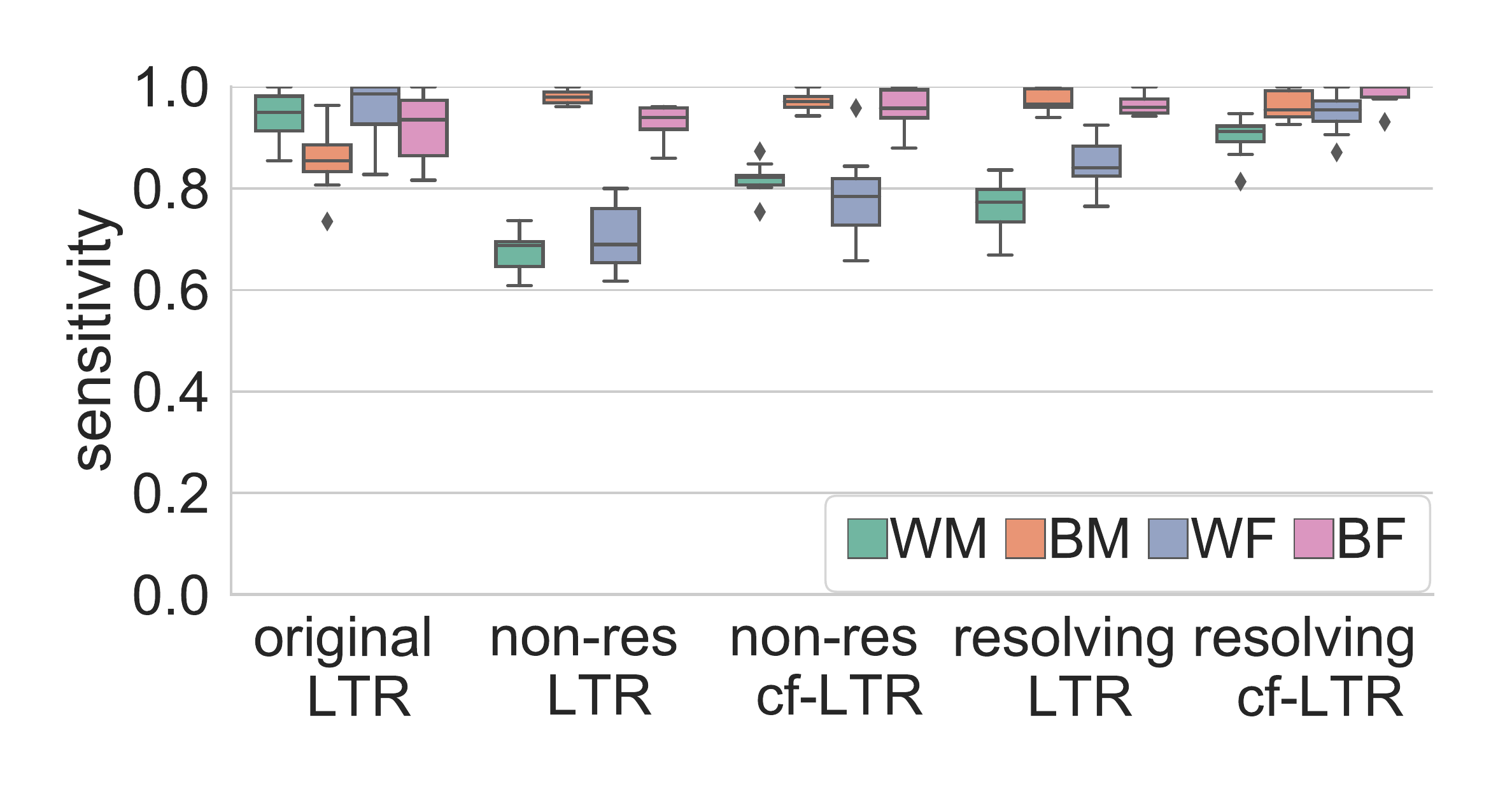}
	\label{fig:cm_eo_ltr}
	}
	\caption{Equal opportunity on \MV and \CM. Treatments: trained and tested on unmodified rankings (original LTR); trained on counterfactually fair rankings with $X$ as non-resolving, unmodified test (non-resolving LTR);  non-resolving training \& test (non-resolving cf-LTR); resolving training \& unmodified test (resolving LTR); non-resolving training \& test (resolving cf-LTR).} 
	\label{fig:eo_ltr}
\end{figure*}

We now investigate the usefulness of our methods for supervised learning of counterfactually fair ranking models.  We use ListNet, a popular LTR algorithm, as implemented by Ranklib~\cite{dang2013lemur}. ListNet is a listwise method --- it takes ranked lists as input and generates predictions in the form of ranked lists. All experiments in this section are executed on 10 pairs of training/test datasets for \MV, or on 10 training/test splits (70\% training and 30\% test) for \CM. 

We conduct experiments in two regimes that differ in their treatment of the test set (see Implementation in Section~\ref{sec:fair}).  In both, we first fit a causal model ${\mathcal M}$ on the training data,  then update the training data to include counterfactually fair values of the score $Y$ and of any non-resolving mediators $X$, and finally train the ranking model ${\mathcal R}$ (\eg ListNet) on the fair training data.  We now have two options: (1) to run ${\mathcal R}$ on the \emph{unmodified (biased) test data}, called LTR in our experiments, or; (2) to \emph{preprocess test data} using ${\mathcal M}$, updating test with counterfactually fair values for the score $Y$ and for any non-resolving mediators $X$, before passing it on to ${\mathcal R}$, called cf-LTR.

Figure~\ref{fig:eo_ltr} presents  performance of LTR on \MV and \CM, comparing five types of pipelines in terms of equal opportunity (EO, see Section~\ref{sec:exp:methods}).  Observe that original LTR achieves high sensitivity for all intersectional groups, and so can be seen as achieving EO.  However, this treatment does not explicitly model ranked fairness considerations.  We also observe that cf-LTR variants (those where test data is processed) outperform their LTR counterparts in terms of sensitivity and EO.    In the \MV example in Figure~\ref{fig:mv_eo_ltr}, the non-resolving LTR variant leads to lower sensitivity for the male groups, likely because women are selected at higher rates, but the cf-LTR variant leads to better sensitivity for males. For \CM in Figure~\ref{fig:cm_eo_ltr},  sensitivity is highest for Black males in all counterfactual treatments, with cf-LTR variants achieving a better balance  across groups.

\begin{figure*}[t!]
	\centering
	\subfloat[\MV, $k=500$]{\includegraphics[width=0.49\linewidth]{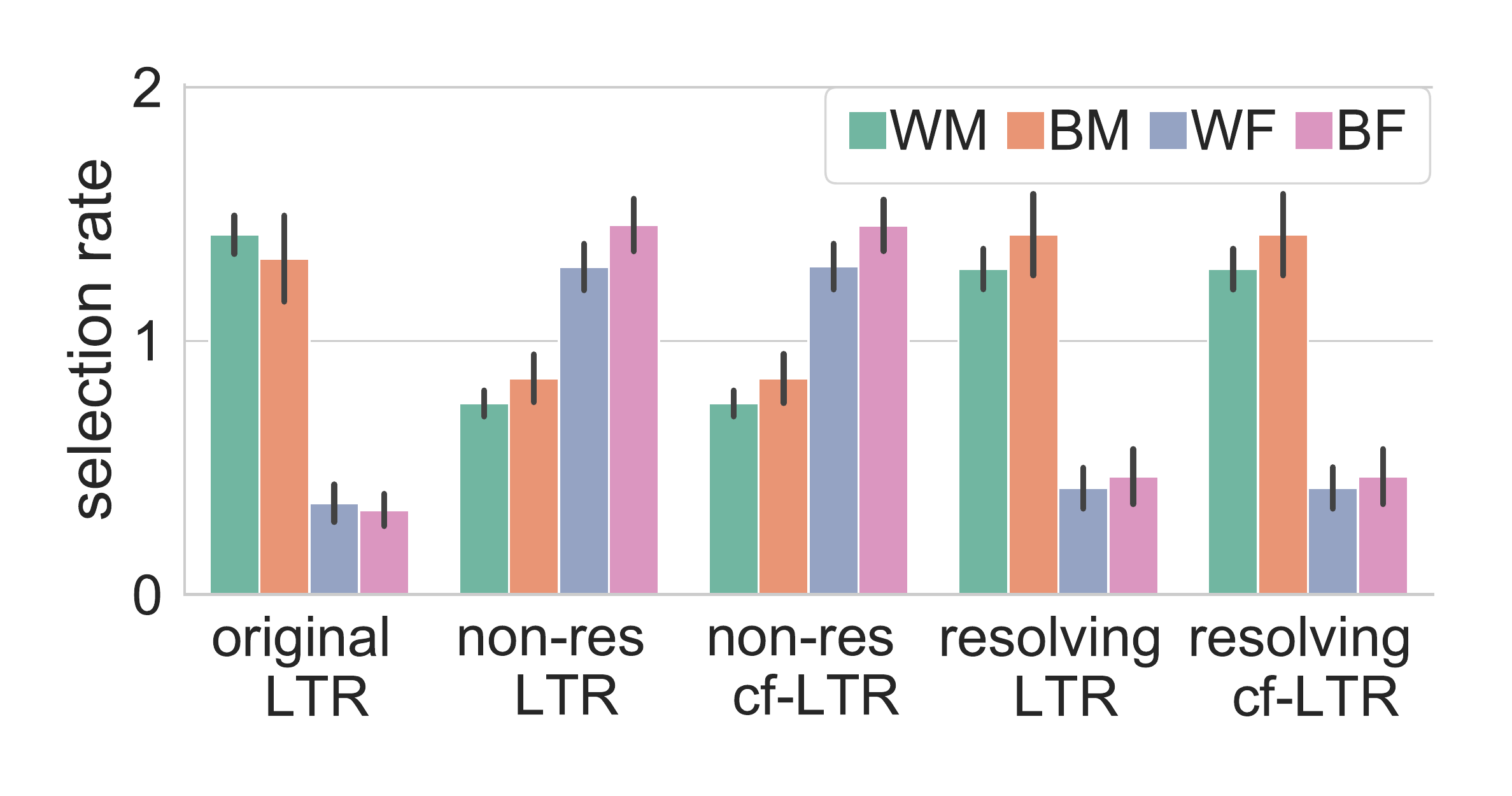}
	\label{fig:mv_dp_ltr}
	}
	\subfloat[\CM, $k=500$]{
	\includegraphics[width=0.49\linewidth]{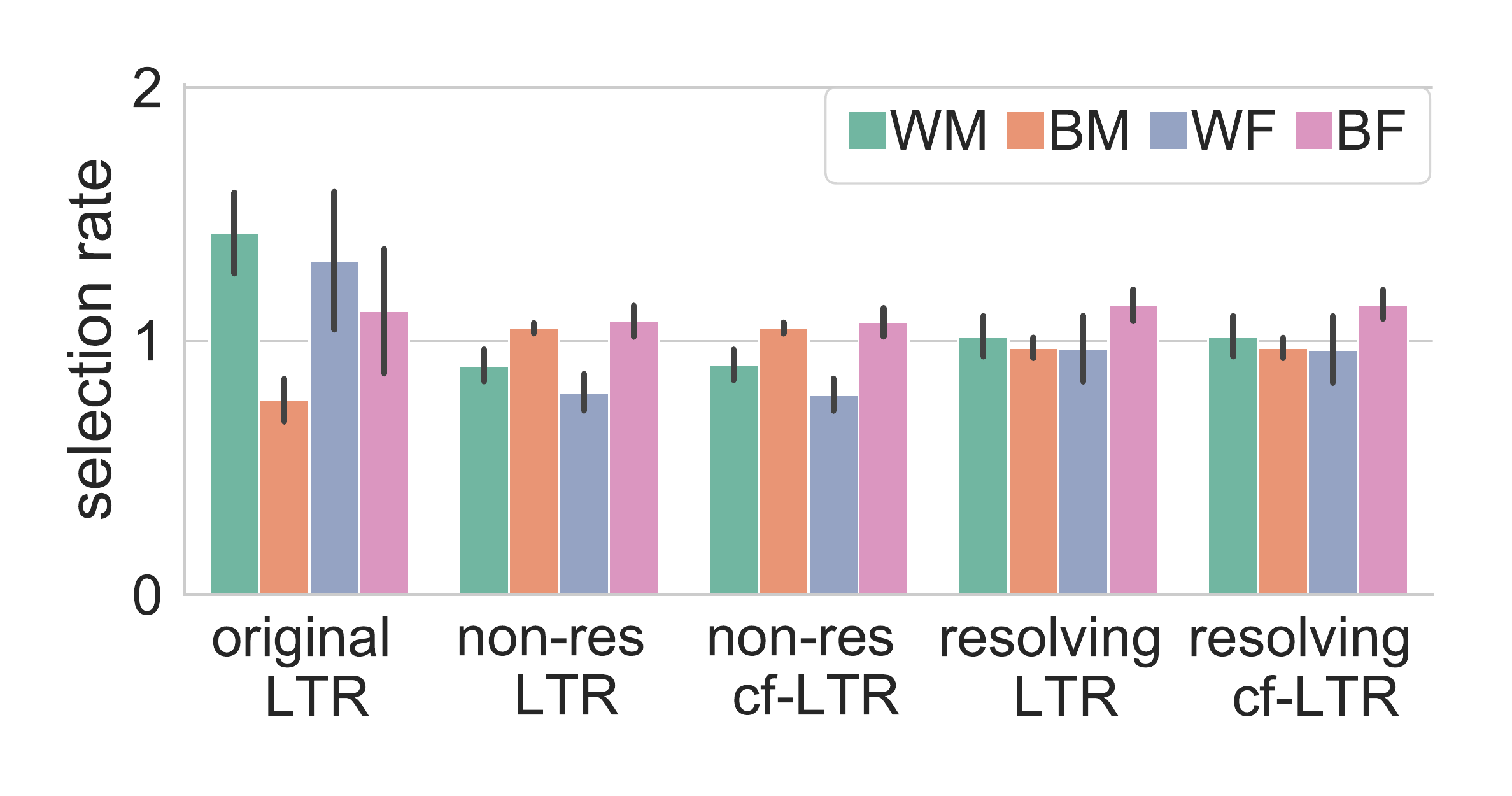}
	\label{fig:cm_dp_ltr}
	}
	\caption{Demographic parity on \MV and \CM. Treatments: trained and tested on unmodified rankings (original LTR); trained on counterfactually fair rankings with $X$ as non-resolving, unmodified test (non-resolving LTR);  non-resolving training \& test (non-resolving cf-LTR); resolving training \& unmodified test (resolving LTR); non-resolving training \& test (resolving cf-LTR).} 
	\label{fig:dp_ltr}
\end{figure*}

Figure~\ref{fig:dp_ltr} shows selection rates for intersectional groups for \MV and \CM. Observe that selection rates for intersectional groups in the test set correspond to what we would expect for these treatments.  For example, \BM are selected at a higher rate for both resolving LTR and resolving cf-LTR than in the original ranking for \MV (Figure~\ref{fig:mv_dp_ltr}).

We also quantified utility as average precision (AP, see Section~\ref{sec:exp:methods}). For \MV, AP is 77\% when unmodified ranking are used for training and test. For counterfactually fair training data with non-resolving $X$ (weight-lifting), AP on unmodified (biased) test is 27\% but it increases to 91\% when test data is preprocessed.  For counterfactually fair training data with resolving $X$, AP is 68\% for unmodified test and 83\% when test is preprocessed.  For \CM, our findings are similar: AP is 63\% on unmodified training and test; when $X$ (number of prior arrests) is non-resolving, AP is 50\%  for LTR, and raises to 74\% for cf-LTR.  When $X$ is resolving, AP is 58\% for LTR and 79\% for cf-LTR. These results show that training on counterfactually fair rankings induce a utility cost, but that this cost can be effectively controlled if there is an opportunity to preprocess test data.  

\section{Related Work}
\label{sec:related}

\paragraph{ Intersectionality} From the seeds of earlier work~\cite{collective1983combahee}, including examples that motivated our experiments~\cite{crenshaw1989demarginalizing}, intersectional feminism has developed into a rich interdisciplinary framework to analyze power and oppression in social relations~\cite{collins2002black, shields2008gender}. We refer especially to~\cite{d2020data, noble2018algorithms} in the context of data and information technology. Other recent technical work in this area focuses on achieving guarantees across intersectional subgroups~\cite{hebert2018multicalibration, DBLP:conf/icml/KearnsNRW18, DBLP:conf/aies/KimGZ19}, including on computer vision tasks~\cite{buolamwini2018gender}, or makes connections to privacy~\cite{DBLP:journals/corr/abs-1807-08362}. These do not take a causal approach or deal with ranking tasks. In our framework intersectionality does not just refer to a redefinition of multiple categorical sensitive attributes into a single product category or inclusion of interaction terms. Specific problems may imply different constraints or interpretations for different sensitives attributes, as shown in the \textit{moving company} example where a mediator is caused by one sensitive attribute but not the other.

\paragraph{Causality and fairness} A growing literature on causal models for fair machine learning~\cite{chiappa2019path, kilbertus2017avoiding, DBLP:conf/nips/KusnerLRS17, nabi2018fair, zhang2018fairness} emphasizes that fairness is a normative goal that relates to real world (causal) relationships. One contribution of the present work is to connect intersectionality and fair ranking tasks to this literature, and therefore to the rich literature on causal modeling. Some recent work in causal fairness focuses on the impact of learning optimal, fair policies, potentially under relaxations of standard causal assumptions that allow interference~\cite{kusner2019making, nabi2019learning}.

\paragraph{Ranking and fairness} While the majority of the work on fairness in machine learning focuses on classification or risk prediction, there is also a growing body of work on fairness and diversity in ranking~\cite{DBLP:conf/sigmod/AsudehJS019,DBLP:conf/icalp/CelisSV18,DBLP:conf/fat/CelisMV20,DBLP:conf/icde/LahotiGW19,DBLP:conf/edbt/StoyanovichYJ18,DBLP:conf/kdd/WuZW18,DBLP:conf/ssdbm/YangS17,DBLP:conf/ijcai/YangGS19,DBLP:conf/www/Zehlike020,DBLP:conf/cikm/ZehlikeB0HMB17}.  Of these, only Yang et al.~\cite{DBLP:conf/ijcai/YangGS19} consider  intersectional concerns, although not in a causal framework.  The authors observe that when representation constraints are stated on individual attributes, like race and gender, and when the goal is to maximize score-based utility subject to these constraints, then a particular kind of unfairness can arise, namely,  utility loss may be imbalanced across intersectional groups.  In our experiments we observed a small imbalance in utility loss across intersectional groups (1-5\%) and will investigate the conditions under which this happens in future work. Finally, \cite{DBLP:conf/kdd/WuZW18} applies causal modeling to fair ranking but estimates scores from observed ranks, uses causal discovery algorithms to learn an SCM, and does not consider intersectionality, while the present work considers the case when scores are observed and the SCM chosen \emph{a priori}.





\section{Conclusion}
\label{sec:conc}

Our work builds on a growing literature for causal fairness to introduce a modeling framework for intersectionality and apply it to ranking. Experiments show this approach can be flexibly applied to different scenarios, including mediating variables, and the results compare reasonably to intuitive expectations we may have about intersectional fairness for those examples. The flexibility of our approach and its connection to causal methodology makes possible a great deal of future work including exploring robustness of rankings to unmeasured confounding~\cite{DBLP:conf/uai/KilbertusBKWS19} or uncertainty about the underlying causal model~\cite{russell2017worlds}. Future technical work can relax some assumptions under specific combinations of model structures, estimation methods, and learning task algorithms. For example, we have shown in experiments that the LTR task (without in-processing) with ListNet works reasonably well, but future work could identify the conditions when this insensitivity of a learned ranker $\hat \btau$ to counterfactual transformations on the training data guarantees counterfactual fairness holds at test time, perhaps with explicit bounds on discrepancies due to issues like covariate shift. 

Like any approach based on causality, our method relies on strong assumptions that are untestable in general, though they may be falsified in specific cases. Sequential ignorability in particular is a stronger assumption in cases with more mediating variables, or with a mediator that is causally influenced by many other variables (observed or unobserved). Such cases increase the number of opportunities for sequential ignorability to be violated for one of the mediators or by one of the many causes of a heavily influenced mediator.

Intersectional fairness is not a purely statistical or algorithmic issue.  As such, any technical method will require assumptions at least as strong as the causal assumptions we make. In particular, there are normative and subtle empirical issues embedded in any approach to fairness, such as the social construction of sensitive attributes, or the choice of which mediators may be considered resolving in our framework. For these reasons \emph{we believe the burden of proof should fall on any approaches starting from assumptions that minimize fairness interventions}, for example, by designating mediators as resolving variables.

\section{Broader Impact}
\label{sec:bi}

This work is focused on mitigating the potential negative impacts of ranking systems on people due to sensitive attributes that may be out of their control. As such, the broader impact of the work is a central focus. But there are many concerns that may arise in the application of our method which we try to anticipate and discuss here.

There are objections to modeling sensitive attributes as causes rather than considering them to be immutable, defining traits. Some of these objections and responses to them are discussed in~\cite{loftus2018causal}. In the present work we proceed with an understanding that the model is a simplified and reductive approximation, and support for deploying an algorithm and claiming it is fair should require an inclusive vetting process where formal models and technical definitions such as these are \emph{tools for inclusively achieving consensus} and not for rubber stamping or obfuscation.

There are many issues outside the scope of the present work but which are important in any real application. Choices of which attributes are sensitive, which mediators are resolving (and for which sensitive attributes), the social construction and definitions of sensitive attributes, choices of outcome/utility or proxies thereof, technical limitations in causal modeling, the potential for (adversarial) misuse are all issues that may have adverse impacts when using our method. We do stress that these are not limitations inherent to our approach in particular, rather, these concerns arise for virtually any approach in a sensitive application. For a high level introductions to these issues, including a causal approach to them, see~\cite{barocas2017fairness,kusner2020long}.


\section{Acknowledgments}
\label{sec:ack}

This research was supported in part by NSF Grants No. 1926250 and 1934464.

\bibliographystyle{abbrvnat}
\bibliography{intersectionality.bib}

\end{document}